\documentclass[10pt,journal,compsoc]{IEEEtran}
\usepackage{helvet}
\usepackage{courier}
\usepackage{graphicx}
\usepackage{amsmath,amsthm,amssymb}
\usepackage{textcomp,booktabs}
\usepackage{footmisc}
\usepackage[normalem]{ulem}
\usepackage{multirow}
\usepackage{setspace}
\usepackage{bigstrut}
\usepackage{array}
\usepackage{amsfonts}
\usepackage{bm}
\usepackage{url}
\usepackage{balance}
\usepackage{mathrsfs}
\usepackage{adjustbox}
\usepackage{caption}
\usepackage{cuted}
\usepackage{capt-of}
\usepackage{xcolor, color, colortbl}
\usepackage{amssymb}
\usepackage{rotating}
\usepackage{commath}
\usepackage[ruled,vlined]{algorithm2e}
\usepackage{algorithmic}
\usepackage{dsfont}
\usepackage{bbding}
\usepackage{threeparttable}
\usepackage{listings}
\usepackage{subfig}

\ifCLASSOPTIONcompsoc
  \usepackage[nocompress]{cite}
\else
  \usepackage{cite}
\fi

\ifCLASSINFOpdf
\else
\fi

\IEEEpubid{
\fbox{
\parbox{0.975\textwidth}{
\copyright 2023 IEEE. Personal use of this material is permitted. Permission from IEEE must be obtained for all other uses, in any current or future media, including\hfill
reprinting/republishing this material for advertising or promotional purposes, creating new collective works, for resale or redistribution to servers or lists, or reuse of\hfill
any copyrighted component of this work in other works. \vspace{-0.5mm}
}%
}}

\newcommand{\Rb}{\mathbb{R}}
\newcommand{\Lm}{\mathcal{L}}
\newcommand{\loss}[1]{$\mathcal{L}_{#1}$}
\newcommand{\hparam}[1]{$\lambda_{#1}$}
\newcommand{\cl}{cl}
\newcommand{\reg}{reg}
\newcommand{\ssl}{ssl}
\newcommand{\ce}{ce}

\newcommand{\Image}{I}
\newcommand{\Label}{Y}
\newcommand{\EMALabe}{\bar{Y}}
\newcommand{\Feat}{F}
\newcommand{\EMAFeat}{\bar{F}}
\newcommand{\Prob}{P}
\newcommand{\EMAProb}{\bar{P}}
\newcommand{\Bank}{\mathcal{B}}
\newcommand{\INBank}{\mathcal{B}}
\newcommand{\Proto}{\mathcal{P}}

\newcommand{\Encoder}{$\Theta_e$\xspace}
\newcommand{\Seghead}{$\Theta_c$\xspace}
\newcommand{\Projhead}{$\Theta_p$\xspace}
\newcommand{\EMAEncoder}{$\Theta'_e$\xspace}
\newcommand{\EMASeghead}{$\Theta'_c$\xspace}
\newcommand{\EMAProjhead}{$\Theta'_p$\xspace}

\newcommand{\etal}{\textit{et al.}}

\newcommand{\fullmethod}{\textbf{Se}mantic-Guided \textbf{Pi}xel \textbf{Co}ntrast\xspace}
\newcommand{\method}{SePiCo\xspace}

\newcommand{\VspaceBefore}{\vspace{-2mm}}
\newcommand{\VspaceAfter}{\vspace{-2mm}}

\definecolor{Gray}{gray}{0.9}

\begin{document}


\title{SePiCo: Semantic-Guided Pixel Contrast for \\Domain Adaptive Semantic Segmentation}


\author{Binhui Xie, Shuang Li, Mingjia Li, Chi Harold Liu,~\IEEEmembership{Senior Member,~IEEE}, Gao Huang, and Guoren Wang
\IEEEcompsocitemizethanks{ 
\IEEEcompsocthanksitem B. Xie, S. Li, M. Li, C. H. Liu and G. Wang are with the School of Computer Science and Technology, Beijing Institute of Technology, Beijing, China. Email: \{binhuixie, shuangli, mingjiali, chiliu, wanggrbit\}@bit.edu.cn. \protect
\IEEEcompsocthanksitem G. Huang is with Department of Automation, Tsinghua University, Beijing, China, Email: gaohuang@tsinghua.edu.cn. \protect
\IEEEcompsocthanksitem Corresponding author: Shuang Li. \protect
% \IEEEcompsocthanksitem X. Cheng and R. Yang are with Inceptio Technology, Shanghai, China. Email: cnorbot@gmail.com, ryang@cs.uky.edu \protect\\
}
}

% \markboth{Journal of \LaTeX\ Class Files,~Vol.~14, No.~8, AUGUST~2015}%
\markboth{IEEE TRANSACTIONS ON PATTERN ANALYSIS AND MACHINE INTELLIGENCE}%
{Shell \MakeLowercase{\textit{et al.}}: Bare Advanced Demo of IEEEtran.cls for IEEE Computer Society Journals}

\IEEEtitleabstractindextext{%
\begin{abstract}
  Domain adaptive semantic segmentation attempts to make satisfactory dense predictions on an unlabeled target domain by utilizing the supervised model trained on a labeled source domain. One popular solution is self-training, which retrains the model with pseudo labels on target instances. Plenty of approaches tend to alleviate noisy pseudo labels, however, they ignore the intrinsic connection of the training data, i.e., intra-class compactness and inter-class dispersion between pixel representations across and within domains. In consequence, they struggle to handle cross-domain semantic variations and fail to build a well-structured embedding space, leading to less discrimination and poor generalization. In this work, we propose \textit{\textbf{Se}mantic-Guided \textbf{Pi}xel \textbf{Co}ntrast (\method)}, a novel one-stage adaptation framework that highlights the semantic concepts of individual pixels to promote learning of class-discriminative and class-balanced pixel representations across domains, eventually boosting the performance of self-training methods. Specifically, to explore proper semantic concepts, we first investigate a \textit{centroid-aware pixel contrast} that employs the category centroids of the entire source domain or a single source image to guide the learning of discriminative features. Considering the possible lack of category diversity in semantic concepts, we then blaze a trail of distributional perspective to involve a sufficient quantity of instances, namely \textit{distribution-aware pixel contrast}, in which we approximate the true distribution of each semantic category from the statistics of labeled source data. Moreover, such an optimization objective can derive a closed-form upper bound by implicitly involving an infinite number of (dis)similar pairs, making it computationally efficient. Extensive experiments show that \method not only helps stabilize training but also yields discriminative representations, making significant progress on both synthetic-to-real and daytime-to-nighttime adaptation scenarios. The code and models are available at https://github.com/BIT-DA/\method.
\end{abstract}

\begin{IEEEkeywords}
  Domain adaptation, semantic segmentation, semantic variations, representation learning, self-training.
\end{IEEEkeywords}}

\maketitle

% is not selected <OR> if conference mode is selected - because compsoc
% conference papers position the abstract like regular (non-compsoc)
\IEEEdisplaynontitleabstractindextext

% \fi
\IEEEpeerreviewmaketitle

% main body
% intro
\ifCLASSOPTIONcompsoc
\IEEEraisesectionheading{\section{Introduction}}
\else
\section{Introduction}
\label{sec:introduction}
\fi

\IEEEPARstart{G}{eneralizing} deep neural networks to an unseen domain is pivotal to a broad range of critical applications such as autonomous driving~\cite{JanaiGBG20,geiger2012autonomous} and medical analysis~\cite{ronneberger2015UNet,HesamianJHK19}. For example, autonomous cars are required to operate smoothly in diverse weather and illumination conditions, e.g., foggy, rainy, snowy, dusty, and nighttime. While humans excel at such scene understanding problems, it is struggling for machines to forecast. Semantic segmentation is a fundamental task relevant that assigns a unique label to every single pixel in the image. Recently, deep Convolution Neural Networks (CNNs) have made rapid progress with remarkable generalization ability~\cite{alexnet,long2015fully,yu2016dilated,chen2018deeplab}. CNNs, however, are quite data-hungry and the pixel-level labeling process is expensive and labor-intensive, thereby restricting their real-world utility. As a trade-off, training with freely-available synthetic data rendered from game engines~\cite{stephan2016gtav,ros2016synthia} turns into a promising alternative. This is not the case, unfortunately, deep models trained on simulated data often drop largely in realistic scenarios due to \textit{domain shift}~\cite{dataset_shift_in_ML09}.

%##################################################################################################
\begin{figure}[t] \centering
    \includegraphics[width=0.42\textwidth]{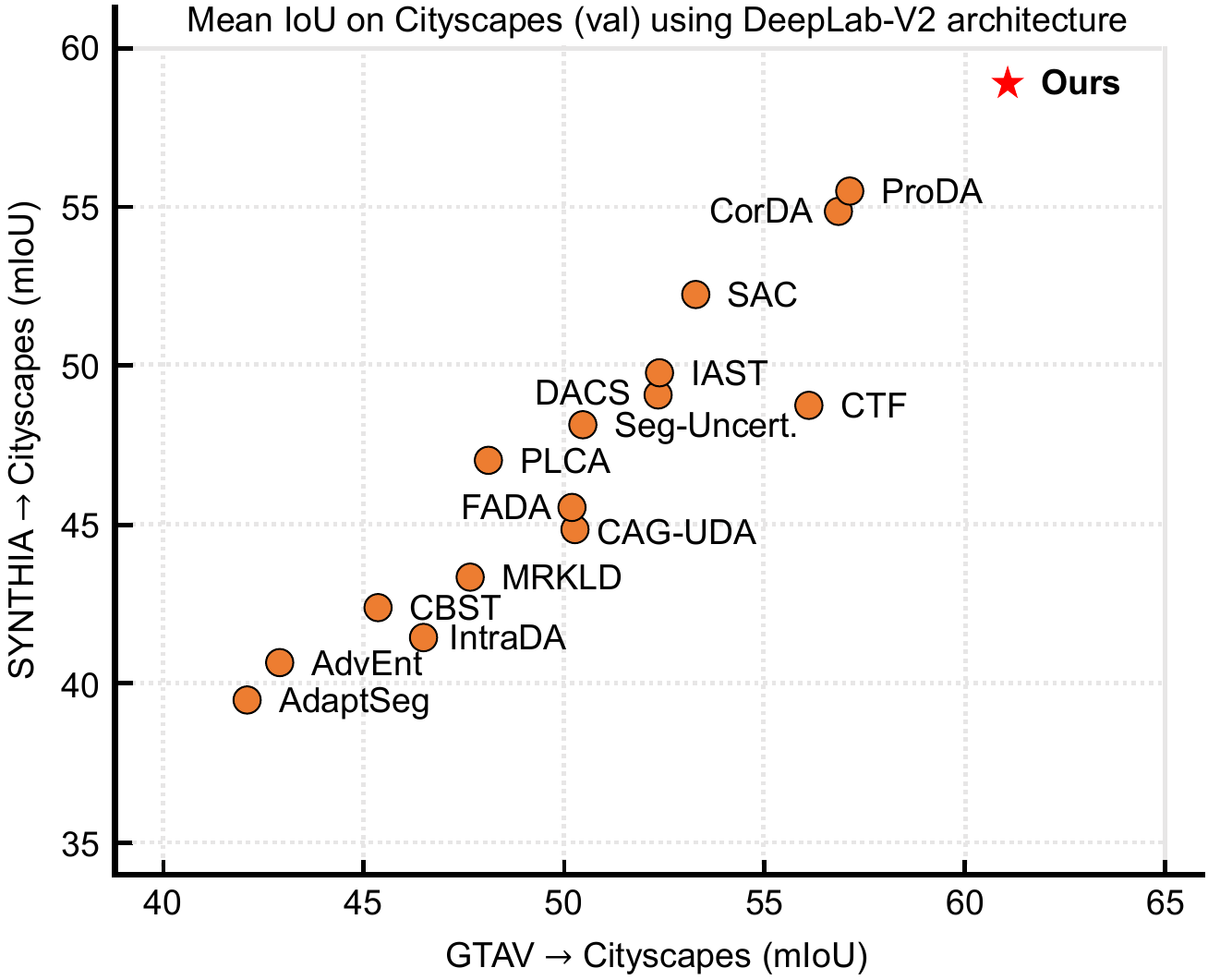} \VspaceBefore
    \caption{Results preview on two popular synthetic-to-real semantic segmentation tasks. Our method is shown in \textbf{bold}.} \VspaceAfter
    \label{Fig_results}
\end{figure}
%##################################################################################################

%##################################################################################################
\begin{figure*}[t] \centering
    \includegraphics[width=0.9\textwidth]{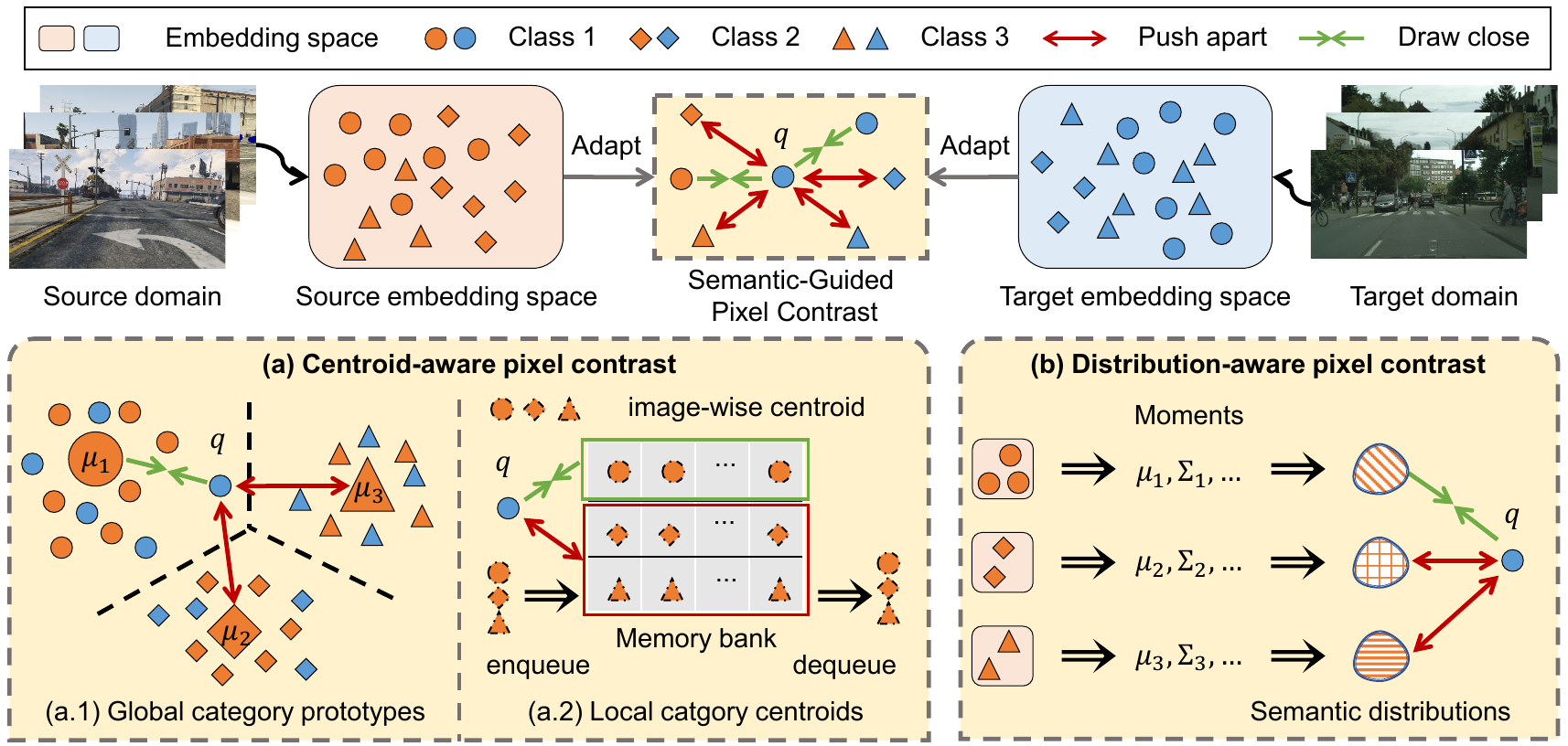} \VspaceBefore
    \caption{{\bf Illustration of the main idea}. By contrastively matching a pixel query $q$ to distinct semantics, features with the same semantic concepts are drawn closer while those with different ones are pushed apart across domains. We first explore (a) \textbf{Centroid-aware pixel contrast} including (a.1) \textbf{global category prototypes} simply computed on the entire source domain, which render the overall appearance of each category and (a.2) \textbf{local category centroids} of each class in a single source image, which are stored into a memory bank. Further, we develop (b) \textbf{Distribution-aware pixel contrast}: the distributions of each category on source features are depicted as class-specific holistic concepts to guide the semantic alignment. 
    } \VspaceAfter
    \label{Fig_motivation}
\end{figure*}
%##################################################################################################

Recent trends of domain adaptation (DA) inspire the emergence of extensive works to transfer knowledge from a label-rich source (synthetic) domain to a label-scarce target (real) domain, which enjoys tremendous success~\cite{pan2010survey,tzeng2017adversarial,long2015dan_tpami,GDCAN,ganin2015dann_jmlr}. Most previous works develop adversarial training algorithms to diminish the domain shift existing in input~\cite{dundar2020stylization,yang2020fda}, feature~\cite{Hoffman_cycada2017,tsai2018learning} or output~\cite{vu2019advent,pan2020unsupervised} space. Despite the fact that the above methods can draw two domains closer globally, it does not guarantee those feature representations from different classes in the target domain are well-separated. Utilizing category information can refine such alignment~\cite{luo2021category,TSA,wang2020differential,wang2020class,zhang2019category}. But, pixels in different images might share much similar semantics while their visual characteristics, such as color, scale, illumination, etc. could be quite different, which is deleterious to the continual learning of pixel representations across two domains.

Another line of work harnesses self-training to promote the segmentation performance~\cite{lukas2021daformer,Dong2021where,DACS_2021_WACV,zheng_2021_IJCV}. By adopting confidence estimation~\cite{Charles2021confidence_estimation,zou2019confidence}, consistency regularization~\cite{SAC_2021_CVPR,Ma_2021_CVPR}, or label denoising~\cite{ProDA_2021_CVPR,Wang_2021_ICCV}, the noisy in pseudo labels could be relieved to some extent. While many works are already capable of establishing milestone performance, there is still much room for improvement beyond the current state-of-the-art. We find that most approaches do not explicitly address the domain discrepancy, and the learned target representations are still dispersed. In addition, many works opt for a stage-wise training mechanism to avoid training error amplification in a single-stage model, which heavily relies on a well-initialized model to increase the reliability of generated pseudo labels. Hereafter, several methods combine adversarial training and self-training~\cite{li2019bidirectional,Ma_2021_CVPR} or train with auxiliary tasks~\cite{Vu_DADA_2019,CorDA_2021_ICCV} to learn discriminative representations from unlabeled target data. 

Contrastive learning is a relevant topic, which learns proper visual representations by comparing different unlabeled data~\cite{he2020momentum,chen2020contrastive,xie2020contrastive_dense,cai2020jcl}.  Without any supervision, models are capable of finding patterns like similarity and dissimilarity. The huge success of contrastive learning and the aforementioned drawbacks in prior arts together motivate us to rethink the current de facto training paradigm in semantic segmentation under a domain shift. Basically, the power of contrastive learning roots in instance discrimination, which takes advantage of semantic concepts within data. With this insight, we find a new path to build models that are robust to distribution shifts by exploring cross-domain pixel contrast under the guidance of proper semantic concepts, which attracts similar pixels and dispels dissimilar ones in a latent space, as illustrated in Fig.~\ref{Fig_motivation}.

In this work, we present a novel end-to-end framework, \method, for domain adaptive semantic segmentation. Not only does \method outperform previous works (Fig.~\ref{Fig_results}), but it is also simple yet effective, keeping one-stage training complexity. Precisely, we build upon a self-training method~\cite{DACS_2021_WACV} and introduce several dense contrastive learning mechanisms. The core is to explore suitable semantic concepts to guide the learning of a well-structured pixel embedding space across domains. Here, a plain way is to adopt the averaged feature of a category over the entire source domain as its global prototype. A prototype could render the overall appearance of a category but might omit variations in some attributes (e.g., shape, color, illumination) of the category, impairing the discriminability of the learned features. To enhance diversity, a natural extension is to enlarge the number of contrastive pairs. We then investigate a memory bank mechanism, in which the averaged features of each category in the current source image are enqueued into a dictionary and the oldest ones are dequeued. Unfortunately, class biases may exist in this mechanism since those under-represented classes (e.g., truck, bus, rider) are updated more slowly. Meanwhile, it is computationally expensive as well.

Grounded on the above discussions, we hypothesize that  if every dimension in the embedding space follows a distribution, pixel representations from a similar semantic class would have a similar distribution, which is independent of the domain. Thereby, we take the distribution of each category in the source domain as a richer and more comprehensive semantic description. The real distribution can be properly estimated with sufficient supervision of source data. This formulation enables a wide variety of samples from estimated distributions, which is tailored for pixel representation learning in dense prediction tasks. Furthermore, we analyze Pixel-wise Discrimination Distance (PDD) to certify the validity of our method regarding pixel-wise category alignment. Extensive experiments demonstrate that contrastively driving the source and target pixel representations towards proper semantic concepts can lead to more effective domain alignment and significantly improve the generalization capacity of the model. We hope this exploration will shed light on future studies.

In a nutshell, our contributions can be summarized:
\begin{itemize}
    \item We provide a new impetus to mitigate domain shift by explicitly enhancing the similarity of pixel features with corresponding semantic concepts and increasing the discrimination power on mismatched pairs, no matter the source or target domain. 
    \item To facilitate efficiency and effectiveness, a closed-form upper bound of the expected contrastive loss is derived with the moments of each category. \method is also a one-stage adaptation framework robust to both daytime and nighttime segmentation situations.
    \item Extensive experiments on popular semantic segmentation benchmarks show that \method achieves superior performance. Particularly, we obtain mIoUs of 61.0\%, 58.1\%, and 45.4\% on benchmarks GTAV $\to$ Cityscapes, SYNTHIA $\to$ Cityscapes, and Cityscapes $\to$ Dark Zurich respectively. Equipped with the latest Transformer, \method further improves by mIoUs of 9.3\%, 5.6\%, and 8.0\% respectively, setting the new state of the arts. Ablation study and throughout analysis verify the effectiveness of each component.
    \item Our \method, aiming at a general framework, can further well generalize to unseen target domains and be effortlessly employed to object detection tasks.
\end{itemize}
%%%%%%%%%%%%%%%%%%%%%%%%%%%%%%%%%%%%%%%%%%%%%%%%%%%%%%%%%%%%%%%%%%%%%%%%%%%%%%%%%%%%%%%%%%%%%%%%%%%

%%%%%%%%%%%%%%%%%%%%%%%%%%%%%%%%%%%%%%%%%%%%%%%%%%%%%%%%%%%%%%%%%%%%%%%%%%%%%%%%%%%%%%%%%%%%%%%%%%%
% related
\section{Related Work}
\label{sec:relatedwork}
Our work draws upon existing literature on semantic image segmentation, domain adaptation, and representation learning. For brevity, we only discuss the most relevant works.

\subsection{Semantic Segmentation}
The recent renaissance in semantic segmentation began with the fully convolutional networks~\cite{long2015fully}. Mainstream methods strive to enlarge receptive fields and capture context information~\cite{yu2016dilated,chen2018deeplab,zhao2017pspnet}. Among them, the family of DeepLab enjoys remarkable popularity because of its effectiveness. Inspired by the success of the Transformers~\cite{Vaswani2017transformer} in natural language processing, many works adopt it to visual tasks including image classification~\cite{Dosovitskiy2021vit} and semantic segmentation~\cite{xie2021segformer}, offering breakthrough performance. These studies, though impressive, require a large amount of labeled datasets and struggle to generalize to new domains.

In this work, we operate semantic segmentation under such a domain shift with the aim of learning an adequate model on the unlabeled target domain. Concretely, we map pixel representations in different semantic classes to a distinctive feature space via a pixel-level contrastive learning formulation. The learned pixel features are not only discriminative for segmentation within the source domain, but also, more critically, well-aligned for cross-domain segmentation.

\subsection{Nighttime Semantic Segmentation}
Nighttime Semantic Segmentation is much more challenging in safe autonomous driving due to poor illuminations and arduous human annotations. Only a handful of works have been investigated in the past few years. Dai \etal~\cite{dai2018dark} introduce a two-step adaptation method with the aid of an intermediate twilight domain. Sakaridis \etal~\cite{sakaridis2020mapguided} leverage geometry information to refine predictions and transfer the style of nighttime images to that of daytime images to reduce the domain gap. 
Recently, Wu \etal~\cite{WU_2021_CVPR} jointly train a translation model and a segmentation model in one stage, which efficiently performs on par with prior methods. 

While daytime and nighttime image segmentation tasks differ only in appearance, current works focus on designing specialized methods for each task. Different from the above methods, \method is able to address both daytime and nighttime image segmentation tasks in a universal framework.

\subsection{Domain Adaptation in Semantic Segmentation}
Domain Adaptation (DA) has been investigated for decades in theory~\cite{ben2007analysis,MDD} and in various tasks~\cite{xie2022active,ganin2015dann_jmlr,ChuDK17,Chen0SDG18,lv2021pareto}. Given the power of DCNNs, deep DA methods have been gaining momentum to significantly boost the transfer performance of a segmentation model. A multitude of works generally fall into two categories: \textit{adversarial training}~\cite{tsai2018learning,dundar2020stylization,wang2020class,luo2021category} and \textit{self-training}~\cite{Cheng_2021_ICCV,DACS_2021_WACV,melas2021pixmatch,ProDA_2021_CVPR}.

\textbf{Adversarial training} methods diminish the distribution shift of two domains at image~\cite{Hoffman_cycada2017,dundar2020stylization,yang2020fda}, feature~\cite{kim2020learning,pan2020unsupervised}, or output~\cite{tsai2018learning,vu2019advent,luo2021category} level in an adversarial manner. To name a few, Hoffman \etal~\cite{Hoffman_cycada2017} bring DA to segmentation by building generative images for alignment. On the other hand, Tsai \etal~\cite{tsai2018learning} suggest that performing alignment in the output space is more practical. A few works also leverage different techniques via entropy~\cite{vu2019advent,pan2020unsupervised} and information bottleneck~\cite{luo2021category}. Other concurrent works~\cite{luo2021category,wang2020class} incorporate category information into the adversarial loss to intensify local semantic consistency. Due to the absence of holistic information about each category, adversarial training is usually less stable. Therefore, some methods instead adopt category anchors~\cite{zhang2019category,wang2020differential,Ma_2021_CVPR} computed on source data to advance the alignment. A recent work~\cite{huang2021category} presents a category contrast method to learn discriminative representation. By contrast, we endeavor to explore semantic concepts from multiple perspectives. More importantly, we set forth a generic semantic-guided pixel contrast to emphasize pixel-wise discriminative learning, allowing us to minimize the intra-class discrepancy and maximize the inter-class margin of pixel representations across domains.

{\bf Self-training} methods exploit unlabeled target data via training with pseudo labels~\cite{MeiZZZ20,SAC_2021_CVPR,Dong2021where,Charles2021confidence_estimation,kang2020pixel}. 
In an example, Zou \etal~\cite{zou2018unsupervised} propose an iterative learning strategy with class balance and spatial prior for target instances. In~\cite{DACS_2021_WACV}, Tranheden \etal~propose a domain-mixed self-training pipeline to avoid training instabilities, which mixes images from two domains along with source ground-truth labels and target pseudo labels. Later on, Wang \etal~\cite{CorDA_2021_ICCV} enhance self-training via leveraging the auxiliary supervision from depth estimation to diminish the domain gap. Lately, Zhang \etal~\cite{ProDA_2021_CVPR} utilize the feature distribution from prototypes to refine target pseudo labels and distill knowledge from a strongly pre-trained model. However, most existing methods always encounter an obstacle in that target representations are dispersed due to the discrepancy across domains. In addition, most of them utilize a warm-up model to generate initial pseudo labels, which is hard to tune. Differently, our framework  performs one-stage end-to-end adaptation produce without using any separate pre-processing stages. In addition, \method can largely improve self-training and easily optimize pixel embedding space.

\subsection{Representation Learning}
To date, unsupervised representation learning has been extensively investigated due to its promising ability to learn representations in the absence of human supervision, especially for contrastive learning~\cite{hadsell2006dimensionality,oord2018infoNCE,he2020momentum,chen2020contrastive,cai2020jcl}. Let $f$ be an embedding function that transforms a sample $x$ to an embedding vector $q=f(x)\,, q \in \mathbb{R}^d$ and let $(x\,,x^+)$ be similar pairs and $(x\,,x^-)$ be dissimilar pairs. Then, normalize $q$ onto a unit sphere and  a popular contrastive loss such as InfoNCE~\cite{oord2018infoNCE} is formulated as:
\begin{small}
    \begin{align}
        \mathbb{E}_{q\,,q^+\,,\{q^{-}_n\}_{n=1}^N} \left[- \log \frac{e^{q^{\top}q^+/\tau}}{e^{q^{\top}q^+/\tau} + \sum_{n=1}^N e^{q^{\top}q^{-}_n/\tau}}\right]. \nonumber
    \end{align}
\end{small}%%
In practice, the expectation is replaced by the empirical estimate. As shown above, the contrastive loss is essentially based on the softmax formulation with a temperature $\tau$. 

Intuitively, the above methods encourage instance discrimination. Recent works~\cite{xie2020contrastive_dense,wang2020contrastive_dense,wenguan_exploring_ICCV,alonso2021semi} also extend contrastive learning to dense prediction tasks. These methods either engage in better visual pre-training for dense prediction tasks~\cite{xie2020contrastive_dense,wang2020contrastive_dense} or explore dense representation learning in the fully supervised setting~\cite{wenguan_exploring_ICCV} or semi-supervised setting~\cite{alonso2021semi}.  Thereby, they generally tend to learn the pixel correspondence on the category of objects that appear in different views of an image rather than learning the semantic concepts across datasets/domains, so the learned representations cannot directly deploy under domain shift. On the contrary, we draw inspiration from contrastive learning and construct contrastive pairs according to different ways of semantic information to bridge the domain shift, which has received limited consideration in the existing literature.
%%%%%%%%%%%%%%%%%%%%%%%%%%%%%%%%%%%%%%%%%%%%%%%%%%%%%%%%%%%%%%%%%%%%%%%%%%%%%%%%%%%%%%%%%%%%%%%%%%%

%%%%%%%%%%%%%%%%%%%%%%%%%%%%%%%%%%%%%%%%%%%%%%%%%%%%%%%%%%%%%%%%%%%%%%%%%%%%%%%%%%%%%%%%%%%%%%%%%%%
% method
\section{Methodology}
\label{sec:method}
In this section, we first briefly introduce the background and illustrate the overall idea in Section~\ref{sec:background}. Then the details of semantic statistics and our framework are elaborated in Section~\ref{sec:sematin_statistics} and Section~\ref{sec:contrastive_adaptation}, respectively. Finally, we present the training procedure and the \method algorithm in Section~\ref{sec:objective}. 

\subsection{Background}
\label{sec:background}
\subsubsection{Problem formulation}
\begin{figure}[t] \centering
    \includegraphics[width=0.48\textwidth]{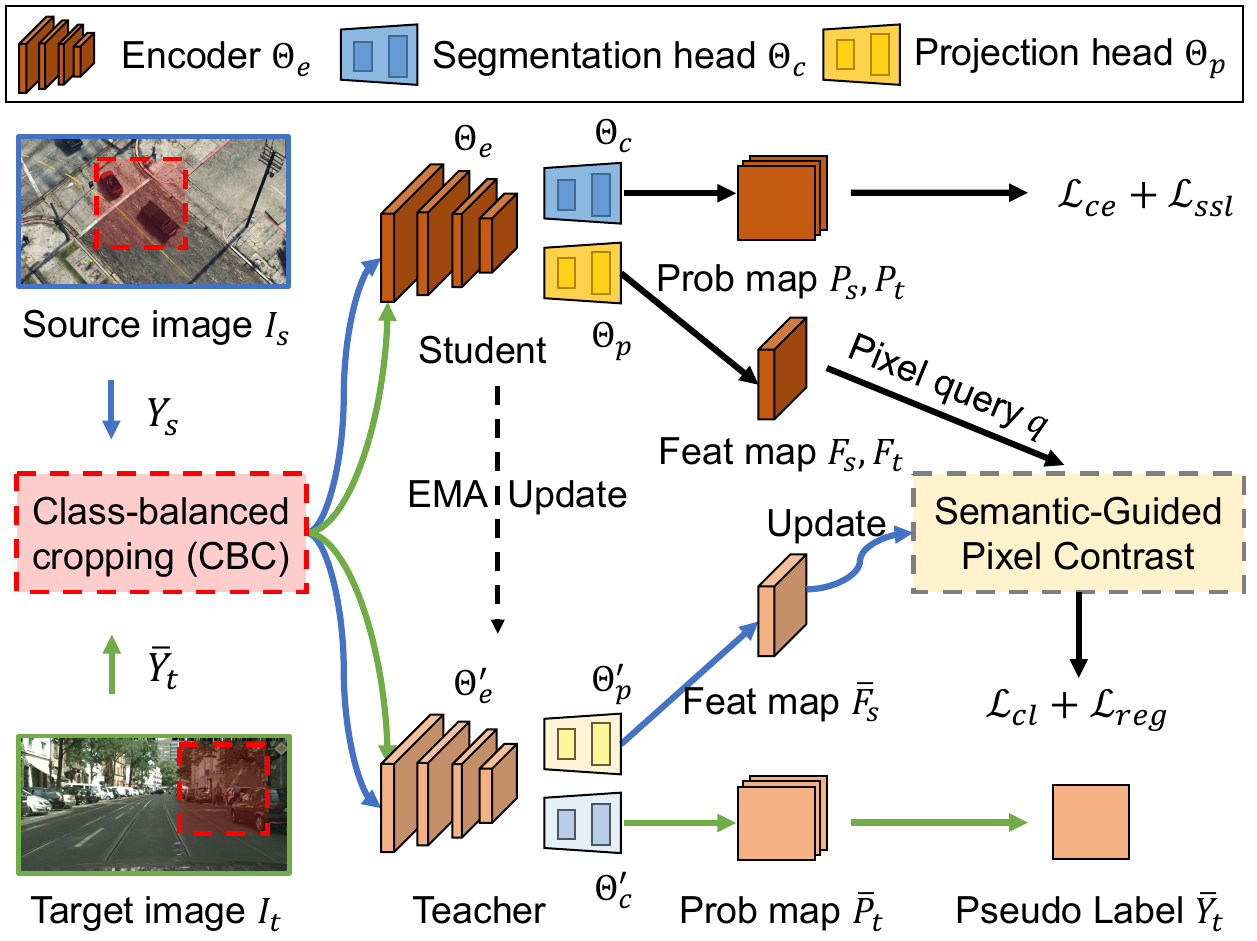} %\VspaceBefore
    \caption{{\bf Framework overview.} First, our basic framework is based on a teacher-student architecture and the teacher model provides source feature map $\EMAFeat_s$ and target pseudo labels $\EMALabe_t$. Second, we propose class-balanced cropping to frequently crop image patches with under-represented objects that balance performance across classes. And third, except for self training losses, $\Lm_{ce}$ and $\Lm_{ssl}$, we contrastively enforce the pixel representations $\Feat_s\,,\Feat_t$ towards centroid-aware or distribution-aware semantics using $\Lm_{cl}$ and $\Lm_{reg}$. After training is completed, we throw away projection head \Projhead and use encoder \Encoder and head \Seghead for segmentation task.} \VspaceAfter
    \label{Fig_framework}
\end{figure}

For domain adaptive semantic segmentation, we have a collection of labeled source data $\Image_s$ with pixel-level labels $\Label_s$ as well as unlabeled target data $\Image_t$. The goal is to categorize each pixel of a target image into one of the predefined $K$ categories through learning a model consisting of a feature encoder \Encoder, a multi-class segmentation head \Seghead, and an auxiliary projection head \Projhead. We adopt the teacher-student architecture~\cite{TarvainenV17} (Teacher networks are denoted as \EMAEncoder, \EMASeghead, and \EMAProjhead.) as our basic framework, shown in Fig.~\ref{Fig_framework}.

During training, images from source and target domains $\Image_s\,, \Image_t\in \Rb^{H\times W\times 3}$ are randomly sampled and passed into both teacher and student networks, respectively. The hidden-layer features $\Feat_s\,, \Feat_t\in \Rb^{H'\times W'\times A}$, and final pixel-level predictions $\Prob_s\,, \Prob_t\in \Rb^{H\times W\times K}$ are generated from the student, where $A$ is the channel dimension of intermediate features and $H' (\ll H)\,, W' (\ll W)$ are spatial dimensions of features. Similarly, we access corresponding source features $\EMAFeat_s\in \Rb^{H'\times W'\times A}$, and target pixel-level predictions $\EMAProb_t$ from the momentum-updated teacher. Note that no gradients will be back-propagated into the teacher network~\cite{TarvainenV17}.

%##################################################################################################
%##################################################################################################

\subsubsection{Self-training domain adaptation revisit}
\label{sec:self_training}
Here, we give an overview of a self-training method~\cite{DACS_2021_WACV} for evaluating different semantic-guided pixel contrasts. Traditional self-training methods usually consider two aspects. On the one hand, these methods train a model \Seghead$\circ $ \Encoder to minimize the categorical cross-entropy (CE) loss in the source domain, formalized as a fully supervised problem:
\begin{small}
    \begin{align}
        \Lm_{ce} = -\frac{1}{HW}\sum_{i\in \{1\,,2\,,\cdots\,, H\times W\}} \sum_{k} \mathds{1}_{[\Label_{s,i} = k]} \log P_{s,i}^{k}\,,
        \label{eq:source_ce_loss}
    \end{align}%
\end{small}%%
where $\Label_{s,i}$ is the one-hot label for pixel $i$ in $\Image_s$ and $\mathds{1}_{[\cdot]}$ is an indicator function that returns 1 if the condition holds or 0 otherwise. On the other hand, to better transfer the knowledge from the source domain to the target domain, self-training usually uses a teacher network to produce more reliable pseudo labels $\EMALabe_{t}$ for a target image,
\begin{small}
    \begin{align}
        \EMALabe_{t,j} = \mathop{\arg\max}\limits_{k} \EMAProb_{t,j}^k\,,~j\in \{1\,,2\,,\cdots\,, H\times W\}\,.
        \label{eq:target_pseudo}
    \end{align}   
\end{small}%
In practice, we compute the pseudo labels online during the training and avoid any additional inference step, which is simpler and more efficient. Specifically, we forward a target image and obtain the pseudo labels using Eq.~\eqref{eq:target_pseudo}. Besides, since the pseudo labels are usually noisy, a confidence estimation is made for generated pseudo labels.  Specifically, the number of pixels with the maximum softmax probability exceeding a threshold $\alpha$ is calculated first: 
\begin{small}
    \begin{align}
        NUM_{conf} = \sum_{j\in \{1\,,2\,,\cdots\,, H\times W\}} \mathds{1}_{[max_{k}\EMAProb_{t,j}^{k} > \alpha]}\,.
        \label{eq:target_pseudo_weight}
    \end{align}   
\end{small}%
Next, the ratio of pixels exceeding the threshold over the whole image serves as confidence weights, $w=\frac{NUM_{conf}}{HW}$ and the student network is re-trained on target data,
\begin{small}
    \begin{align}
        \Lm_{ssl} = -\frac{1}{HW}\sum_{j\in \{1\,,2\,,\cdots\,, H\times W\}} \sum_{k} w \cdot \mathds{1}_{[\EMALabe_{t,j} = k]} \log P_{t,j}^{k}\,.
        \label{eq:target_ce_loss}
    \end{align}   
\end{small}%
Finally, let's go back to the teacher network. The weights of teacher network \EMAEncoder, \EMASeghead, \EMAProjhead are set as the exponential moving average (EMA) of the weights of student network \Encoder, \Seghead, \Projhead in each iteration~\cite{TarvainenV17}. Take \EMAEncoder as an example,
\begin{small}
    \begin{align}
        \Theta'_e \leftarrow \beta \Theta'_e + (1-\beta) \Theta_e\,,
        \label{eq:ema_update}
    \end{align}   
\end{small}%
where $\beta$ is a momentum parameter. Similarly, \EMASeghead, \EMAProjhead should also be updated via Eq.~\eqref{eq:ema_update}. In this work, $\beta$ is fixed to 0.999.

Note that incorporating data augmentation with self-training has been shown to be particularly efficient~\cite{SAC_2021_CVPR,DACS_2021_WACV}. Following~\cite{DACS_2021_WACV}, we use the teacher network to generate a set of pseudo labels $\EMALabe_t$ on the weakly-augmented target data. Concurrently, the student network is trained on strongly-augmented target data. We use standard resize and random flip as the weak augmentation. Strong augmentation includes color jitter, gaussian blur, and ClassMix~\cite{OlssonTPS21}.

\subsubsection{Overall motivation}
As mentioned before, however, there is a major limitation of traditional self-training methods: most of them neglect explicit domain alignment. As a result, even under perfect pseudo labeling on target samples, the negative transfer may exist, causing pixel features from different domains but of the same semantic class to be mapped farther away. To sidestep this issue, we promote semantic-guided representation learning in the embedding space. A naive way is to directly adopt global category prototypes computed on the source domain to guide the alignment between source and target domains. An obstacle to this design, however, is that prototypes could only reflect the common characteristic of each category but cannot fully unlock the potential strength of semantic information, leading to erroneous representation learning. Inspired by~\cite{he2020momentum}, we go a step further and store the local image centroids of each source image into a memory bank so that the semantic information exploited is roughly proportional to the size of the bank. But this mechanism will arise class bias as features of some classes, e.g., bicycles, pedestrians and poles, rarely appear. Meanwhile, this does consume a lot of computing resources.

Consequently, to promote diversity in semantic concepts, we newly introduce the distribution-aware pixel constant to contrastively strengthen the connections between each pixel representation and estimated distributions. Moreover, such a distribution-aware mechanism could be viewed as training on infinite data and is more computation efficient, which is intractable for a memory bank.

\subsection{Semantic Statistics Calculation}
\label{sec:sematin_statistics}
Given the source feature map $\EMAFeat_s \in \Rb^{H'\times W'\times A}$ from the teacher model, for any pixel indexed $i\in \{1\,,2\,,\cdots\,, H'\times W'\}$ in $\EMAFeat_s$, we first divide its feature into the set of $k^{th}$ semantic class, i.e., $\Lambda^k$ according to its mask $M_{s, i} \in \Rb^{H'\times W'}$ downsampled from ground truth label. Hereafter, the local centroid of the $k^{th}$ category in an image is calculated by
\begin{small}
    \begin{align}
        \mu'^k = \frac{1}{|\Lambda^k|} \sum_{i\in \{1\,,2\,,\cdots\,, H'\times W'\}} \mathds{1}_{[M_{s, i}=k]} \EMAFeat_{s, i}\,,
        \label{eq:local_image_centroid}
    \end{align}
\end{small}%
where $|\cdot|$ is the cardinality of the set.

For centroid-aware semantic information, we require either global category prototypes or local category centroids. On the one side, we opt for an online fashion on the entire source domain, aggregating mean statistics one by one to build global category prototypes. Mathematically, the online estimate algorithm for the mean of the $k^{th}$ category is given by
 \begin{small}
     \begin{align}
         \mu^k_{(t)} = \frac{n^k_{(t-1)}\mu^k_{(t-1)} + m^k_{(t)} {\mu'}^k_{(t)}}{n^k_{(t-1)} + m^k_{(t)}}\,,
         \label{eq:mu}
     \end{align}
 \end{small}%
where $n^k_{(t-1)}$ is the total number of pixels belonging to the $k^{th}$ category in previous $t-1$ images, and $m^k_{(t)}$ is the number of pixels belonging to the $k^{th}$ category in current $t^{th}$ image. Thereby, we are allowed to obtain $K$ global prototypes:
 \begin{small}
     \begin{align}
        \Proto=\{\mu^1\,, \mu^2\,, \cdots\,, \mu^K\} \,.
     \end{align}
 \end{small}%
 On the other side, we maintain local centroids of each class from the latest source images to form a dynamic categorical dictionary with $K$-group queue, 
\begin{small}
    \begin{align}
        \Bank=\{\INBank^1\,,\INBank^2\,,\cdots\,,\INBank^K\} \,,
    \end{align}
\end{small}%
where $\INBank^k=\{\mu'^k_{(t-B)}\,, \mu'^k_{(t-B+1)}\,, \cdots\,, \mu'^k_{(t)}\}$. $B$ is the shared queue size for all queues. Note that the oldest centroids are dequeued and currently computed centroids are enqueued.

 \textbf{Discussion:} \textit{merit and demerit of centroid-aware statistics}. Each global category prototype renders the overall appearance of one category, yet it might omit diversity and impair the discriminability of the learned representations. On the other way, the memory bank is able to expand the set of negative and positive samples, thus it can embrace more semantic information. More importantly, almost all semantic information could be covered when $B$ is large enough, however, it is neither elegant nor efficient in presenting pixel embedding space. To capture and utilize rich semantic information as efficiently and comprehensively as possible, we try to build from the distributional perspective as follows.
 
We observe that pixel representations with respect to each class will have a similar distribution. With this in mind, we propose to build the distribution-aware semantic concepts with sufficient labeled source instances. Therefore, we need to acquire the covariance of the multidimensional feature vector $\EMAFeat_{s,i}$ for a better representation of the variance between any pair of elements in the feature vector. The covariance matrix $\Sigma^k$ for category $k$ can be updated via
\begin{small}
    \begin{align}
        \Sigma^k_{(t)} &= \frac{n^k_{(t-1)}\Sigma^k_{(t-1)}+m^k_{(t)}{\Sigma'}^k_{(t)}}{n^k_{(t-1)}+m^k_{(t)}} \nonumber\\
        &+\frac{n^k_{(t-1)}m^k_{(t)}\left(\mu^k_{(t-1)}-{\mu'}^k_{(t)}\right)\left(\mu^k_{(t-1)}-{\mu'}^k_{(t)}\right)^{\top}}{\left(n^k_{(t-1)} + m^k_{(t)}\right)^2} \,,
        \label{eq:Sigma}
    \end{align}%
\end{small}%
where ${\Sigma'}^k_{(t)}$ is the covariance matrix of the features between the $k^{th}$ category in the $t^{th}$ image. It is noteworthy that $K$ mean vectors and $K$ covariance matrices are initialized to zeros. During training, we dynamically update these statistics using Eq.~\eqref{eq:mu} and Eq.~\eqref{eq:Sigma} with source feature map $\EMAFeat_s$ from momentum-updated teacher network. The estimated distribution-aware semantic statistics are more informative to guide the pixel representation learning between domains. 

\subsection{\fullmethod}
\label{sec:contrastive_adaptation}
In the literature, a handful of methods have leveraged categorical feature centroids~\cite{zhang2019category,Ma_2021_CVPR,huang2021category,wang2020differential} as anchors to remedy domain shift, yielding promising results. However, few attempts have been made in this regime to quantify the distance between features of different categories. It is arduous to separate pixel representations with similar semantic information in target data as no supervision information is available, which severely limits their potential capability in dense prediction tasks. On the contrary, we design a unified framework to integrate three distinct contrastive losses that target learning similar/dissimilar pairs at the pixel level to mitigate the domain gap via either centroid-aware pixel contrast or distribution-aware pixel contrast.

As stated above, the pixel representation separation in the source domain is naturally guaranteed by source mask $M_s$ from the ground truth label. Similarly, for target data, we desire to obtain satisfactory target mask $M_t$ for each pixel via generated pseudo labels from $\EMALabe_t$ (Eq.~\eqref{eq:target_pseudo}). Whereafter, any pixel representation either in source or target feature maps (defined as the pixel query $q\in \Rb^{A}$ for simplicity) now needs to yield a low loss value when simultaneously forming multiple positive pairs $(q\,, q^{+}_m)$ and multiple negative pairs $(q\,, q^{k-}_n)$, where $q^{+}_m$ indicates the $m^{th}$ positive example from the same category considering $q$ and $q^{k-}_n$ represents $n^{th}$ negative example from the $k^{th}$ different class. Formally, we define a new pixel contrast loss function for $q$:
\begin{small}
    \begin{align}
        \ell^{cl}_q = -\frac{1}{M}\sum_{m=1}^M\log\frac{e^{q^{\top} q^{+}_m/\tau}}{e^{q^\top q^{+}_m/\tau} + \sum\limits_{k\in\mathcal{K}^-}\frac{1}{N}\sum\limits_{n=1}^N e^{q^\top q^{k-}_n/\tau}}\,,
    \end{align}
\end{small}%%
where $M$ and $N$ are the numbers of positive and negative pairs and $\mathcal{K}^-$ denotes the set containing all different classes from that of $q$. In the following sections, we will describe three pixel contrast losses $\ell^{protocl}_q$, $\ell^{bankcl}_q$ and $\ell^{distcl}_q$ respectively to derive a better-structured embedding space, eventually boosting the performance of segmentation model. 

In short, we enable learning discriminative pixel representations across domains via a unified contrastive loss
\begin{small}
    \begin{align}
        \mathcal{L}_{cl} = \frac{1}{|\Psi|}\sum_{q\in \Feat_s \cup \Feat_t} \ell^{cl}_{q} \,, \label{eq:loss_cl}
    \end{align}% 
\end{small}%
where $|\Psi|$ is the total number of pixels in the union of $F_s$ and $F_t$. Note that such contrastive loss is employed in both domains simultaneously. For one thing, when the loss is applied in the source domain, the student network is capable of yielding more discriminative representations for pixel-level predictions, which increases the robustness of the model. Another effect is that the target representations are contrastively adapted in a pixel-wise manner, which benefits minimizing the intra-category discrepancy and maximizing the inter-category margin and facilitates transferring knowledge from source to target explicitly.

Moreover, except for individual pixel representation learning, we introduce a regularization term to make the feature representations of input images globally diverse and smooth, which is formalized as 
\begin{small}
    \begin{align}
        \mathcal{L}_{reg} = \frac{1}{K\log K} \sum_{k=1}^{K} \log\frac{e^{Q^{\top} \mu^{k}/\tau}}{\sum_{l=1}^{K} e^{Q^\top \mu^{l}/\tau}}\,,
    \end{align}
\end{small}%
where $Q=\frac{1}{H'\times W'} \sum_{i\in\{1\,,2\,,\cdots\,,H'\times W'\}}\Feat_{s/t,i}$ is the mean feature representation of a source or target image. This objective is similar to the diversity-promoting objective used in prior DA methods~\cite{LiangHF20}, but is employed in the embedding space. It could circumvent the trivial solution where all unlabeled target data have the same feature encoding.

\subsubsection{Centroid-aware Pixel Contrast}
\label{sec:centroid_adaptation}
Here, we introduce two variants of centroid-aware pixel contrast, namely \method (ProtoCL) and \method (BankCL).

CASE 1: \textbf{ProtoCL} ($M=N=1$). Naively operate $K$ global category prototypes to establish one positive pair and $K-1$ negative pairs. We consider this formulation as the prototype pixel contrast loss function
\begin{small}
    \begin{align}
        \ell^{protocl}_q = -\log\frac{e^{q^{\top} \mu^{+}/\tau}}{e^{q^\top \mu^{+}/\tau} + \sum\limits_{k\in\mathcal{K}^-} e^{q^\top \mu^{k-}/\tau}}\,,
        \label{eq:protocl}
    \end{align}
\end{small}%%
where $\mu^{+}$ is the positive prototype belonging to the same category as the specific query $q$ and $\mu^{k-}$ is the prototype of the $k^{th}$ different category.

CASE 2: \textbf{BankCL} ($M=N=B$). To involve more negative and positive samples for representation learning, we could access more contrastive pairs from a memory bank, in which local category centroids of a single source image are stored. We consider this formulation as the bank pixel contrast loss function
\begin{small}
    \begin{align}
        \ell^{bankcl}_q = -\mathbb{E}_{q^{+}\in \Bank^{+}}\log\frac{e^{q^{\top} q^{+}/\tau}}{e^{q^\top q^{+}/\tau} + \sum\limits_{k\in\mathcal{K}^-} \mathbb{E}_{q^{k-}\in \Bank^{k-}} e^{q^\top q^{k-}/\tau}}\,,
        \label{eq:bankcl}
    \end{align}
\end{small}%%
where $\Bank^{+}$ is the queue comprised of positive samples and $\Bank^{k-}$ refs to  a queue containing negative ones. In Section~\ref{sec:ablation}, we will analyze the effect of bank size $B$.

\textbf{Discussion:} \textit{merit and demerit of centroid-aware pixel contrast}. 
In a word, global prototypes or local centroids can be used as good contrastive samples to pull similar pixel representations closer and push those dissimilar pixel representations away in the embedding space. However, from Eq.~\eqref{eq:protocl} and Eq.~\eqref{eq:bankcl}, we can theorize that the main difference between them is the number of positive and negative pairs. Because of this, if the number of contrastive pairs does matter, it is intuitively reasonable that an infinite number of such pairs would contribute to the establishment of a more robust and discriminative embedding space. We will justify this assumption from the distributional perspective.

\subsubsection{Distribution-aware Pixel Contrast}
\label{sec:distribution_adaptation}

In this part, we derive a particular form of contrastive loss where infinite positive/negative pixel pairs are simultaneously involved with regard to each pixel representation in the source and target domain. A naive implementation is to explicitly sample $M$ examples from the estimated distribution that has the same latent class and $N$ examples from each of the other distributions featuring different semantic concepts. Unfortunately, this is not computationally feasible when $M$ and $N$ are large, as carrying all positive/negative pairs in an iteration would quickly drain the GPU memory. 

To get around this issue, we take an infinity limit on the number of $M$ and $N$, where the effect of $M$ and $N$ are hopefully absorbed in a probabilistic way. With this application of infinity limit, the statistics of the data are sufficient to achieve the same goal of multiple pairing. As $M\,,N$ goes to infinity, it becomes the estimation of:
\begin{small}
    \begin{align}
        &\ell^{\infty}_q = \lim_{M\to \infty\,,N\to \infty} \ell^{cl}_q \nonumber \\ 
        & = \lim\limits_{M\to \infty} -\frac{1}{M}\sum_{m=1}^M \log\frac{e^{q^\top q^+/\tau}}{e^{q^\top q^+/\tau} + \sum\limits_{k\in\mathcal{K}^-} \lim\limits_{N\to \infty} \frac{1}{N}\sum_{n=1}^N e^{q^\top q^{k-}_{n}/\tau}}   \nonumber\\
        & = -\mathbb{E}_{q^{+} \sim p(q^+)} \log\frac{e^{q^\top q^+/\tau}}{e^{q^\top q^+/\tau} + \sum\limits_{k\in\mathcal{K}^-} \mathbb{E}_{q^{k-}\sim p(q^{k-})} e^{q^\top q^{k-}/\tau}} \,, \nonumber
    \end{align}
\end{small}%%
where $p(q^+)$ is the positive semantic distribution with the same label as $q$ and $p(q^{k-})$ is the $k^{th}$ negative semantic distribution with a different label from that of each query $q$. The analytic form of the above is intractable, but it has a rigorous closed form of upper bound, which can be derived
\begin{small}
    \begin{align}
    &\ell^{\infty}_q = -\mathbb{E}_{q^{+}} \log\frac{e^{q^\top q^+/\tau}}{e^{q^\top q^+/\tau} + \sum\limits_{k\in\mathcal{K}^-}\mathbb{E}_{q^{k-}} e^{q^\top q^{k-}/\tau}} \nonumber\\
    &\leq \log\left[ \mathbb{E}_{q^+}\left[e^{\frac{q^\top q^+}{\tau}}+\sum\limits_{k\in\mathcal{K}^-} \mathbb{E}_{q^{k-}}e^{\frac{q^\top q^{k-}}{\tau}}\right]\right] - q^\top\mathbb{E}_{q^+}\left[ \frac{q^+}{\tau}\right] \nonumber\\
    &= \log\left[\mathbb{E}_{q^+}e^{\frac{q^\top q^+}{\tau}} + \sum\limits_{k\in\mathcal{K}^-} \mathbb{E}_{q^{k-}}e^{\frac{q^\top q^{k-}}{\tau}}\right] - q^\top\mathbb{E}_{q^+}\left[\frac{q^+}{\tau}\right]  \nonumber \\
    &= \ell^{distcl}_q  \,, \label{eq:upper_bound}
    \end{align}
\end{small}%%
where the above inequality follows from Jensen's inequality on concave functions, i.e., $\mathbb{E} \log(X) \leq \log \mathbb{E}(X)$. Thus, distribution-aware pixel contrast loss, i.e., \method (DistCL) is yielded to implicitly explore infinite samples.

Next, to facilitate our formulation, we further need an assumption on the feature distribution. For any random variable $x$ that follows Gaussian distribution $x\sim \mathcal{N}(\mu, \Sigma)$, we have the moment generation function~\cite{wang2021isda} that satisfies:
\begin{small}
    \begin{align}
        \mathbb{E} \left[e^{a^\top x}\right] = e^{a^\top \mu + \frac{1}{2}a^\top \Sigma a}\,, \nonumber
    \end{align}%
\end{small}%
where $\mu$ is the expectation of $x$, $\Sigma$ is the covariance matrix of $x$. Therefore, we assume that $q^{+} \sim \mathcal{N}(\mu^{+}\,, \Sigma^{+})$ and $q^{k-}\sim \mathcal{N}(\mu^{k-}\,, \Sigma^{k-})$, where $\mu^{+}$ and $\Sigma^{+}$ are respectively the statistics i.e., mean and covariance matrix, of the positive semantic distribution for $q$, $\mu^{k-}$ and $\Sigma^{k-}$ are respectively the statistics of the $k^{th}$ negative distribution.
Under this assumption, Eq.~\eqref{eq:upper_bound} for a certain pixel representation $q$ immediately reduces to
\begin{small}
    \begin{align}
    \ell^{distcl}_q &=\log \left[e^{\frac{q^\top \mu^{+}}{\tau} + \frac{q^{\top}\Sigma^{+} q}{2\tau^2}} + \sum\limits_{k\in\mathcal{K}^-} e^{\frac{q^\top \mu^{k-}}{\tau} + \frac{q^\top \Sigma^{k-} q}{2\tau^2}} \right] - \frac{q^\top\mu^{+}}{\tau} \nonumber \\ 
    &=-\log\frac{e^{\frac{q^\top \mu^{+}}{\tau}+\frac{q^\top \Sigma^{+} q}{2\tau^2}}}{e^{\frac{q^\top\mu^{+}}{\tau}+\frac{q^\top\Sigma^{+} q}{2\tau^2}}+\sum\limits_{k\in\mathcal{K}^-} e^{\frac{q^\top\mu^{k-}}{\tau}+\frac{q^\top\Sigma^{k-} q}{2\tau^2}}} + \frac{q^\top\Sigma^{+}q}{2 \tau^2} \,. \nonumber
    \end{align}
\end{small}%%
Eventually, the overall loss function regarding each pixel-wise representation thereby boils down to the closed form whose gradients can be analytically solved. 

%##################################################################################################
\begin{algorithm}[t] \centering
    \caption{Pseudocode of class-balanced cropping on an unlabeled target image (PyTorch-style)}.
    \label{alg:cbc_code}
    \definecolor{codeblue}{rgb}{0.25,0.5,0.5}
    \lstset{
        backgroundcolor=\color{white},
        basicstyle=\fontsize{7.2pt}{7.2pt}\ttfamily\selectfont,
        columns=fullflexible,
        breaklines=true,
        captionpos=b,
        commentstyle=\fontsize{7.2pt}{7.2pt}\color{codeblue},
        keywordstyle=\fontsize{7.2pt}{7.2pt},
    }
    \begin{lstlisting}[language=python]
    # img: an unlabeled target image to be cropped
    # pl: corresponding pseudo label of img
    # cat_max_ratio: max ratio of a category in img
    
    best_score = -1, best_crop_box = None  # initialize

    # randomly crop N_crop times and get the best crop
    for _ in range(N_crop):
        score = 0  # initial score
        # get a random box crop
        box_crop = get_random_box_crop(img) 
        pl_crop = crop(pl, box_crop)  # crop pl by crop_box
        # unique classes with pixel count in cropped pl
        classes, cnt = unique_with_counts(pl_crop)
        # category max ratio should be satisfied
        if max(cnt) / sum(cnt) < cat_max_ratio:
            score = sum(log(cnt))  # calculate score
        # compare and get the best
        if score > best_score:
            best_score, best_box_crop = score, box_crop
    
    # perform class-balanced cropping (CBC)
    img = crop(img, best_box_crop)
    pl = crop(pl, best_box_crop)
    \end{lstlisting}
\end{algorithm}
%##################################################################################################

\subsection{Training Procedure}
\label{sec:objective}
In brief, the training procedure of \method can be optimized in a one-stage manner, and we further introduce class-balanced cropping in Alg.~\ref{alg:cbc_code} to stabilize and regularize the process. We summarize the algorithm in Alg.~\ref{alg:contrastive}.

\subsubsection{Class-balanced Cropping (CBC)}
\label{sec:class_balance_cropping}

As we are all aware, realistic segmentation datasets are highly imbalanced. Thus, one challenge of training a capable model under distribution shift is overfitting to the majority classes of the source domain. One can solve this during training through \textit{class-balanced sampling over the entire dataset} like rare class sampling (RCS) in DAFormer~\cite{lukas2021daformer}. Though this strategy is effective, it is only suited to source-domain data with ground-truth labels. Unfortunately, there are no available annotations for target-domain data. To handle this issue, we utilize generated target pseudo labels and provide an alternative strategy, \textit{class-balanced cropping within a single image}, to crop image regions that jointly promote class balance in pixel number and diversity of internal categories (see Alg.~\ref{alg:cbc_code}\footnote{We fix N\_crop=10 and cat\_max\_ratio=0.75 for all experiments.}). Note that we turn to this online strategy for the fact that pseudo labels are constantly changing, thus collecting class statistics over the whole target dataset (like RCS on the whole source dataset) could be much more inefficient. On this basis, we employ RCS for the entire source domain while CBC for a single target image. Accordingly, samples with smaller class frequencies throughout the source domain will have a higher sampling probability, while regions of an unlabeled target image with multiple classes will enjoy a higher cropping probability. Experimentally, we also compare RCS and CBC in Sec.~\ref{sec:ablation}.

%##################################################################################################
\begin{algorithm}[t]
    \DontPrintSemicolon
    \nl {\bf Input}: Input data $\Image_s, \Label_s, \Image_t$, bank size $B$, parameters \hparam{\cl}\,, \hparam{\reg} and maximum/warm-up iteration $L$/$L_w$. \\
    \nl Initialize \Encoder with ImageNet pre-trained parameters and randomly initialize two heads \Seghead and \Projhead. \\
    \nl Initialize statistics $\{\mu^{k}\}_{k=1}^{K}$ and $\{\Sigma^{k}\}_{k=1}^{K}$ to zeros. \\
    \nl Teachers init: \EMAEncoder $\leftarrow$ \Encoder, \EMASeghead $\leftarrow$ \Seghead, \EMAProjhead $\leftarrow$ \Projhead.\\
    \nl \For{$iter\leftarrow$ 0 to $L$}
    {
        \nl Randomly sample a source image $\Image_s$ with $\Label_s$ and a target image $\Image_{t}$.

        \nl Apply class-balance cropping on both $\Image_s$ and $\Image_t$.
        
        \nl Obtain feature maps $\Feat_s$ and $\Feat_t$ and separate pixel-wise representations in the embedding space using corresponding masks $M_s$ and $M_t$.

        \nl Update mean $\{\mu^{k}\}_{k=1}^{K}$ via Eq.~\eqref{eq:mu} and covariance matrices $\{\Sigma^{k}\}_{k=1}^{K}$ via Eq.~\eqref{eq:Sigma} or memory bank with current image-wise centroids $\{{\mu}'^k\}_{k=1}^{K}$.

        \If{$iter > L_w$}{
            \nl Train $\Theta_e\,,\Theta_c\,,\Theta_p$ using \loss{\ce}\,, \loss{\ssl}\,, \loss{\cl}\,, \loss{\reg}\,.
        }
        \Else{
            \nl Train $\Theta_e\,,\Theta_c$ using \loss{\ce}\,, \loss{\ssl}\,.
        }
        \nl Update $\Theta'_e\,,\Theta'_c\,,\Theta'_p$ with $\Theta_e\,,\Theta_c\,,\Theta_p$ via Eq.~\eqref{eq:ema_update}.
    }
    {\bf Return}: Final network weights \Seghead and \Encoder.
    \caption{{\bf \method algorithm.} 
    \VspaceAfter
    \label{alg:contrastive}}
\end{algorithm} 
%##################################################################################################

\subsubsection{Optimization Objective}
The well-known self-training extensively studied in previous methods~\cite{zhang2019category,zou2019confidence,ProDA_2021_CVPR,Wang_2021_ICCV}, is usually achieved by iteratively generating a set of pseudo labels based on the most confident predictions on target data. Nevertheless, it primarily depends on a good initialization model and is hard to tune. Our \method aims to learn a discriminative embedding space and is complementary to the self-training. Therefore, we unify both into a one-stage, end-to-end pipeline to stabilize training and yield discriminative features, which promotes the generalization ability of the model. The overall training objective is formulated as:
\begin{small}
    \begin{align}
        \mathop{\min}\limits_{\Theta_e,\Theta_c,\Theta_p} \Lm_{ce} + \Lm_{ssl} + \lambda_{cl} \Lm_{cl} + \lambda_{reg} \Lm_{reg}\,,
        \label{eq:overall_loss}
    \end{align}
\end{small}%%
where $\lambda_{cl}\,,\lambda_{reg}$ are constants controlling the strength of corresponding loss. Initial tests suggest that using equal weights to combine the $\Lm_{cl}$ with $\Lm_{reg}$ yields better results. For simplicity, both are set to 1.0 without any tuning. By optimizing Eq.~\eqref{eq:overall_loss}, clusters of pixels belonging to the same category are pulled together in the feature space while synchronously pushed apart from other categories, which eventually establishes a discriminative embedding space. In this way, our method can simultaneously minimize the gap across domains as well as enhance the intra-class compactness and inter-class separability in a unified framework. Meanwhile, it is beneficial for the generation of reliable pseudo labels which in turn facilitates self-training.
%%%%%%%%%%%%%%%%%%%%%%%%%%%%%%%%%%%%%%%%%%%%%%%%%%%%%%%%%%%%%%%%%%%%%%%%%%%%%%%%%%%%%%%%%%%%%%%%%%%

%%%%%%%%%%%%%%%%%%%%%%%%%%%%%%%%%%%%%%%%%%%%%%%%%%%%%%%%%%%%%%%%%%%%%%%%%%%%%%%%%%%%%%%%%%%%%%%%%%%
% experiment

%##################################################################################################
\begin{table*} \centering
    \caption{Comparison results of {\bf GTAV $\to$ Cityscapes}. All methods are based on DeepLab-V2 with ResNet-101 for a fair comparison. The best result is highlighted in {\bf bold}.} \label{table:gta} \VspaceBefore
    \resizebox{\textwidth}{!}{
    \begin{tabular}{ l | c c c c c c c c c c c c c c c c c c c |c }
        \toprule[1.2pt]
        Method & road & side. & buil. & wall & fence & pole & light & sign & veg. & terr. & sky & pers. & rider & car & truck & bus & train & mbike & bike & mIoU \\
        \hline
        Source Only & 70.2 & 14.6 & 71.3 & 24.1 & 15.3 & 25.5 & 32.1 & 13.5 & 82.9 & 25.1 & 78.0 & 56.2 & 33.3 & 76.3 & 26.6 & 29.8 & 12.3 & 28.5 & 18.0 & 38.6 \\
        AdaptSeg~\cite{tsai2018learning} & 86.5 & 36.0 & 79.9 & 23.4 & 23.3 & 23.9 & 35.2 & 14.8 & 83.4 & 33.3 & 75.6 & 58.5 & 27.6 & 73.7 & 32.5 & 35.4 & 3.9 & 30.1 & 28.1 & 42.4 \\
        CLAN~\cite{luo2021category} & 88.7 & 35.5 & 80.3 & 27.5 & 25.0 & 29.3 & 36.4 & 28.1 & 84.5 & 37.0 & 76.6 & 58.4 & 29.7 & 81.2 & 38.8 & 40.9 & 5.6 & 32.9 & 28.8 & 45.5 \\
        CBST~\cite{zou2018unsupervised} & 91.8 & 53.5 & 80.5 & 32.7 & 21.0 & 34.0 & 28.9 & 20.4 & 83.9 & 34.2 & 80.9 & 53.1 & 24.0 & 82.7 & 30.3 & 35.9 & 16.0 & 25.9 & 42.8 & 45.9 \\

        MRKLD~\cite{zou2019confidence} & 91.0 & 55.4 & 80.0 & 33.7 & 21.4 & 37.3 & 32.9 & 24.5 & 85.0 & 34.1 & 80.8 & 57.7 & 24.6 & 84.1 & 27.8 & 30.1 & 26.9 & 26.0 & 42.3 & 47.1 \\
        PLCA~\cite{kang2020pixel} & 84.0 & 30.4 & 82.4 & 35.3 & 24.8 & 32.2 & 36.8 & 24.5 & 85.5 & 37.2 & 78.6 & 66.9 & 32.8 & 85.5 & 40.4 & 48.0 & 8.8 & 29.8 & 41.8 & 47.7 \\
        BDL~\cite{li2019bidirectional} & 91.0 & 44.7 & 84.2 & 34.6 & 27.6 & 30.2 & 36.0 & 36.0 & 85.0 & 43.6 & 83.0 & 58.6 & 31.6 & 83.3 & 35.3 & 49.7 & 3.3 & 28.8 & 35.6 & 48.5 \\
        SIM~\cite{wang2020differential} & 90.6 & 44.7 & 84.8 & 34.3 & 28.7 & 31.6 & 35.0 & 37.6 & 84.7 & 43.3 & 85.3 & 57.0 & 31.5 & 83.8 & 42.6 & 48.5 & 1.9 & 30.4 & 39.0 & 49.2 \\
        CaCo~\cite{huang2021category} & 91.9 & 54.3 & 82.7 & 31.7 & 25.0 & 38.1 & 46.7 & 39.2 & 82.6 & 39.7 & 76.2 & 63.5 & 23.6 & 85.1 & 38.6 & 47.8 & 10.3 & 23.4 & 35.1 & 49.2 \\
        ConDA~\cite{Charles2021confidence_estimation} & 93.5 & 56.9 & 85.3 & 38.6 & 26.1 & 34.3 & 36.9 & 29.9 & 85.3 & 40.6 & 88.3 & 58.1 & 30.3 & 85.8 & 39.8 & 51.0 & 0.0 & 28.9 & 37.8 & 49.9 \\
        FADA~\cite{wang2020class} & 91.0 & 50.6 & 86.0 & 43.4 & 29.8 & 36.8 & 43.4 & 25.0 & 86.8 & 38.3 & 87.4 & 64.0 & 38.0 & 85.2 & 31.6 & 46.1 & 6.5 & 25.4 & 37.1 & 50.1 \\  
        LTIR~\cite{kim2020learning} & 92.9 & 55.0 & 85.3 & 34.2 & 31.1 & 34.9 & 40.7 & 34.0 & 85.2 & 40.1 & 87.1 & 61.0 & 31.1 & 82.5 & 32.3 & 42.9 & 0.3 & 36.4 & 46.1 & 50.2 \\
        CAG-UDA~\cite{zhang2019category} & 90.4 & 51.6 & 83.8 & 34.2 & 27.8 & 38.4 & 25.3 & 48.4 & 85.4 & 38.2 & 78.1 & 58.6 & 34.6 & 84.7 & 21.9 & 42.7 & 41.1 & 29.3 & 37.2 & 50.2 \\
        PixMatch~\cite{melas2021pixmatch} & 91.6 & 51.2 & 84.7 & 37.3 & 29.1 & 24.6 & 31.3 & 37.2 & 86.5 & 44.3 & 85.3 & 62.8 & 22.6 & 87.6 & 38.9 & 52.3 & 0.7 & 37.2 & 50.0 & 50.3 \\
        Seg-Uncert.~\cite{zheng_2021_IJCV} & 90.4 & 31.2 & 85.1 & 36.9 & 25.6 & 37.5 & 48.8 & 48.5 & 85.3 & 34.8 & 81.1 & 64.4 & 36.8 & 86.3 & 34.9 & 52.2 & 1.7 & 29.0 & 44.6 & 50.3 \\
        FDA-MBT~\cite{yang2020fda} & 92.5 & 53.3 & 82.4 & 26.5 & 27.6 & 36.4 & 40.6 & 38.9 & 82.3 & 39.8 & 78.0 & 62.6 & 34.4 & 84.9 & 34.1 & 53.1 &  16.9 & 27.7 & 46.4 &  50.5 \\
        KATPAN~\cite{Dong2021where} & 90.8 & 49.8 & 85.1 & 39.5 & 28.4 & 30.5 & 43.1 & 34.7 & 84.9 & 38.9 & 84.7 & 62.6 & 31.6 & 85.1 & 38.7 & 51.8 & 26.2 & 35.4 & 42.6 & 51.8 \\
        DACS~\cite{DACS_2021_WACV} & 89.9 & 39.7 & 87.9 & 30.7 & 39.5 & 38.5 & 46.4 & 52.8 & 88.0 & 44.0 & 88.8 & 67.2 & 35.8 & 84.5 & 45.7 & 50.2 & 0.0 & 27.3 & 34.0 & 52.1 \\
        MetaCorrection~\cite{MetaCorrection_2021_CVPR} & 92.8 & 58.1 & 86.2 & 39.7 & 33.1 & 36.3 & 42.0 & 38.6 & 85.5 & 37.8 & 87.6 & 62.8 & 31.7 & 84.8 & 35.7 & 50.3 & 2.0 & 36.8 & 48.0 & 52.1 \\
        IAST~\cite{MeiZZZ20} & 94.1 & 58.8 & 85.4 & 39.7 & 29.2 & 25.1 & 43.1 & 34.2 & 84.8 & 34.6 & 88.7 & 62.7 & 30.3 & 87.6 & 42.3 & 50.3 & 24.7 & 35.2 & 40.2 & 52.2 \\
        UPLR~\cite{Wang_2021_ICCV} & 90.5 & 38.7 & 86.5 & 41.1 & 32.9 & 40.5 & 48.2 & 42.1 & 86.5 & 36.8 & 84.2 & 64.5 & 38.1 & 87.2 & 34.8 & 50.4 & 0.2 & 41.8 & 54.6 & 52.6 \\
        DPL-dual~\cite{Cheng_2021_ICCV} & 92.8 & 54.4 & 86.2 & 41.6 & 32.7 & 36.4 & 49.0 & 34.0 & 85.8 & 41.3 & 86.0 & 63.2 & 34.2 & 87.2 & 39.3 & 44.5 & 18.7 & 42.6 & 43.1 & 53.3 \\
        SAC~\cite{SAC_2021_CVPR} & 90.4 & 53.9 & 86.6 & 42.4 & 27.3 & 45.1 & 48.5 & 42.7 & 87.4 & 40.1 & 86.1 & 67.5 & 29.7 & 88.5 & 49.1 & 54.6 & 9.8 & 26.6 & 45.3 & 53.8 \\
        CTF~\cite{Ma_2021_CVPR} & 92.5 & 58.3 & 86.5 & 27.4 & 28.8 & 38.1 & 46.7 & 42.5 & 85.4 & 38.4 & \bf 91.8 & 66.4 & 37.0 & 87.8 & 40.7 & 52.4 & \bf 44.6 & 41.7 & 59.0 & 56.1 \\
        CorDA~\cite{CorDA_2021_ICCV} & 94.7 & 63.1 & 87.6 & 30.7 & 40.6 & 40.2 & 47.8 & 51.6 & 87.6 & \bf 47.0 & 89.7 & 66.7 & 35.9 & 90.2 & 48.9 & 57.5 & 0.0 & 39.8 & 56.0 & 56.6 \\
        ProDA~\cite{ProDA_2021_CVPR} & 87.8 & 56.0 & 79.7 & \bf 46.3 & \bf 44.8 & \bf 45.6 & 53.5 & 53.5 & 88.6 & 45.2 & 82.1 & 70.7 & 39.2 & 88.8 & 45.5 & 59.4 & 1.0 & \bf 48.9 & 56.4 & 57.5 \\
        \hline
        {\bf \method (ProtoCL)} & 95.6 & 69.2 & \bf 89.0 & 40.8 & 38.6 & 44.3 & \bf 56.3 & \bf 64.4 & 88.3 & 46.5 & 88.6 & \bf 73.1 & 47.6 & 90.7 & 58.9 & 53.8 & 5.4 & 22.4 & 43.8 & 58.8 \\
        {\bf \method (BankCL)} & \bf 96.1 & \bf 72.1 & 88.6 & 43.1 & 42.4 & 43.7 & 56.0 & 63.5 & \bf 88.9 & 44.5 & 89.0 & 72.7 & 45.7 & 91.1 & 61.7 & 59.6 & 0.0 & 24.7 & 53.6 & 59.8 \\  
        {\bf \method (DistCL)} & 95.2 & 67.8 & 88.7 & 41.4 & 38.4 & 43.4 & 55.5 & 63.2 & 88.6 & 46.4 & 88.3 & \bf 73.1 & \bf 49.0 & \bf 91.4 & \bf 63.2 & \bf 60.4 & 0.0 & 45.2 & \bf 60.0 & \bf 61.0 \\
        \bottomrule[1.2pt]
    \end{tabular} } \VspaceAfter
\end{table*}
%##################################################################################################

\section{Experiment}
\label{sec:experiment}
In this section, we validate \method on two popular synthetic-to-real tasks and challenging daytime-to-nighttime tasks. 
First, we describe datasets and implementation details. 
Next, numerous experimental results are reported for comparison across diverse datasets and architectures. 
Finally, we conduct detailed analyses to obtain a complete picture.
\subsection{Experimental Setups}

\subsubsection{Datasets}
\noindent{\bf GTAV}~\cite{stephan2016gtav} is a composite image dataset sharing 19 classes with Cityscapes. 24,966 city scene images are extracted from the physically-based rendered computer game ``Grand Theft Auto V'' and are used as source domain data for training. 

\noindent{\bf SYNTHIA}~\cite{ros2016synthia} is a synthetic urban scene dataset. Following~\cite{tsai2018learning,lukas2021daformer}, we select its subset, called SYNTHIA-RAND-CITYSCAPES, that has 16 common semantic annotations with Cityscapes. In total, 9,400 images with the resolution 1280$\times$760 from SYNTHIA dataset are used as source data.

\noindent{\bf Cityscapes}~\cite{Cordts2016Cityscapes} is a dataset of real urban scenes taken from 50 cities in Germany and neighboring countries. We use finely annotated images which consist of 2,975 training images, 500 validation images, and 1,525 test images, with a resolution at 2048$\times$1024. Each pixel of the image is divided into 19 categories. For synthetic-to-real adaptation~\cite{lukas2021daformer,ProDA_2021_CVPR,DACS_2021_WACV}, we adopt training images as unlabeled target domain and operate evaluations on its validation set.  For daytime-to-nighttime adaptation~\cite{WU_2021_CVPR,sakaridis2020mapguided,SakaridisDG19}, we use all images from the training set as the source training data.

\noindent{\bf Dark Zurich}~\cite{SakaridisDG19} is another real-world dataset consisting of 2,416 nighttime images, 2,920 twilight images and 3,041 daytime images, with a resolution of 1920$\times$1080. Following~\cite{WU_2021_CVPR}, we utilize 2,416 day-night image pairs as target training data and another 151 test images as target test data that serves as an online benchmark evaluating via online site\footnote{\url{https://competitions.codalab.org/competitions/23553}}.

\subsubsection{Implementation Details}
\textbf{Network architecture.} Our implementation is based on the mmsegmentation toolbox\footnote{https://github.com/open-mmlab/mmsegmentation}. For CNN-based architectures, we utilize the DeepLab-V2~\cite{chen2018deeplab} with ResNet101~\cite{he2016deep} as the backbone. For recent Transformer-based ones, we adopt the same framework used in DAFormer~\cite{lukas2021daformer} as a strong backbone. As for the segmentation head, We follow the mainstream pipelines~\cite{tsai2018learning,lukas2021daformer,SAC_2021_CVPR,DACS_2021_WACV}. Subsequently, a projection head is integrated into the network that maps high-dimensional pixel embedding into a 512-d $\ell_2$-normalized feature vector~\cite{wenguan_exploring_ICCV}. It consists of two 1$\times$1 convolutional layers with ReLU. For fairness, all backbones are initialized using the weights pre-trained on ImageNet~\cite{deng2009imagenet}, with the remaining layers being initialized randomly.

\noindent{\bf Training.} Our model is implemented in PyTorch~\cite{paszke2019pytorch} and trained on a single NVIDIA Tesla V100 GPU. We use the AdamW~\cite{loshchilov2018decoupled} as our optimizer with betas $(0.9, 0.999)$ and weight decay $0.01$. The learning rate is initially set to $6\times 10^{-5}$ for the encoder and $6\times 10^{-4}$ for decoders. 
Similar to~\cite{lukas2021daformer}, learning rate warmup policy and rare class sampling are also applied. In all experiments, we set trade-offs $\lambda_{\cl}\,,\lambda_{\reg}$ to 1.0, 1.0 and threshold $\alpha$, momentum $\beta$, and bank size $B$ to 0.968, 0.999, 200 respectively. We train the network with a batch of two 640$\times$640 random crops for a total of 40k iterations. The statistics in Section~\ref{sec:sematin_statistics} are estimated right from the beginning, but pixel contrast starts from $L_w$ (default  3k) iteration to stabilize training. 

\noindent{\bf Testing.} At the test stage, we only resize the validation images to 1280$\times$640 as the input image. Note that there is no extra inference step inserted into the basic segmentation model, that is, the teacher network, projection head \Projhead, and memory bank $\Bank$, are directly discarded. We employ per-class intersection-over-union (IoU) and mean IoU over all classes as the evaluation metric which is broadly adopted in semantic segmentation~\cite{WU_2021_CVPR,tsai2018learning,lukas2021daformer,ProDA_2021_CVPR}. 

\subsection{Experimental Results}
We comprehensively compare our \method with the recently leading approaches in two representative synthetic-to-real adaptation scenarios: GTAV $\to$ Cityscapes in TABLE~\ref{table:gta}, and SYNTHIA $\to$ Cityscapes in TABLE~\ref{table:synthetic} and a challenging daytime-to-nighttime scenario: Cityscapes $\to$ Dark Zurich in TABLE~\ref{table:dark_zurich}. Additionally, we provide some qualitative results in Fig.~\ref{Fig_seg_map} and Fig.~\ref{Fig_seg_map_dark}. Next, due to the great potential of Vision Transformer, we evaluate our framework on the above three benchmarks and list results in TABLE~\ref{table:transformer}. Last but not least, \method can be applied to domain generalization setting in TABLE~\ref{tab:night_generalization} and detection task in TABLE~\ref{table:detection}.

%##################################################################################################
\begin{table*} \centering
  \caption{Comparison results of {\bf SYNTHIA $\to$ Cityscapes}. mIoU$^{*}$ denotes the mean IoU of 13 classes excluding the classes with $^{*}$. All methods are based on DeepLab-V2 with ResNet-101 for a fair comparison. The best result is highlighted in {\bf bold}.} \label{table:synthetic} \VspaceBefore
  \resizebox{\textwidth}{!}{
      \begin{tabular}{ l | c c c c c c c c c c c c c c c c | c c}
        \toprule[1.2pt]
        Method & road & side. & buil. & wall$^{*}$ & fence$^{*}$ & pole$^{*}$ & light & sign & veg. & sky & pers. & rider & car & bus & mbike & bike & mIoU & mIoU$^{*}$ \\
        \hline
        Source Only & 55.6 & 23.8 & 74.6 & 9.2 & 0.2 & 24.4 & 6.1 & 12.1 & 74.8 & 79.0 & 55.3 & 19.1 & 39.6 & 23.3 & 13.7 & 25.0 & 33.5 & 38.6 \\ 
        AdaptSeg~\cite{tsai2018learning} & 79.2 & 37.2 & 78.8 & 10.5 & 0.3 & 25.1 & 9.9 & 10.5 & 78.2 & 80.5 & 53.5 & 19.6 & 67.0 & 29.5 & 21.6 & 31.3 & 39.5 & 45.9 \\ 
        CLAN~\cite{luo2021category} & 82.7 & 37.2 & 81.5 & - & - & - & 17.7 & 13.1 & 81.2 & 83.3 & 55.5 & 22.1 & 76.6 & 30.1 & 23.5 & 30.7 & - & 48.8 \\ 
        CBST~\cite{zou2018unsupervised} & 68.0 & 29.9 & 76.3 & 10.8 & 1.4 & 33.9 & 22.8 & 29.5 & 77.6 & 78.3 & 60.6 & 28.3 & 81.6 & 23.5 & 18.8 & 39.8 & 42.6 & 48.9 \\
        LTIR~\cite{kim2020learning} & 92.6 & 53.2 & 79.2 & - & - & - & 1.6 & 7.5 & 78.6 & 84.4 & 52.6 & 20.0 & 82.1 & 34.8 & 14.6 & 39.4 & - & 49.3 \\
        MRKLD~\cite{zou2019confidence} & 67.7 & 32.2 & 73.9 & 10.7 & 1.6 & 37.4 & 22.2 & 31.2 & 80.8 & 80.5 & 60.8 & 29.1 & 82.8 & 25.0 & 19.4 & 45.3 & 43.8 & 50.1 \\
        BDL~\cite{li2019bidirectional} & 86.0 & 46.7 & 80.3 & - & - & - & 14.1 & 11.6 & 79.2 & 81.3 & 54.1 & 27.9 & 73.7 & 42.2 & 25.7 & 45.3 & - & 51.4 \\
        SIM~\cite{wang2020differential} & 83.0 & 44.0 & 80.3 & - &- & - & 17.1 & 15.8 & 80.5 & 81.8 & 59.9 & 33.1 & 70.2 & 37.3 & 28.5 & 45.8 & - & 52.1 \\
        FDA-MBT~\cite{yang2020fda} & 79.3 & 35.0 & 73.2 & - & - &  - &  19.9 &  24.0 & 61.7 & 82.6 &  61.4 &  31.1 &  83.9 & 40.8 & 38.4 & 51.1 & - & 52.5 \\
        CAG-UDA~\cite{zhang2019category} & 84.7 & 40.8 & 81.7 & 7.8 & 0.0 & 35.1 & 13.3 & 22.7 & 84.5 & 77.6 & 64.2 & 27.8 & 80.9 & 19.7 & 22.7 & 48.3 & 44.5 & - \\
        MetaCorrection~\cite{MetaCorrection_2021_CVPR} & 92.6 & 52.7 & 81.3 & 8.9 & 2.4 & 28.1 & 13.0 & 7.3 & 83.5 & 85.0 & 60.1 & 19.7 & 84.8 & 37.2 & 21.5 & 43.9 & 45.1 & 52.5 \\
        FADA~\cite{wang2020class} & 84.5 & 40.1 & 83.1 & 4.8 & 0.0 & 34.3 & 20.1 & 27.2 & 84.8 & 84.0 & 53.5 & 22.6 & 85.4 & 43.7 & 26.8 & 27.8 & 45.2 & 52.5 \\ 
        ConDA~\cite{Charles2021confidence_estimation} & 88.1 & 46.7 & 81.1 & 10.6 & 1.1 & 31.3 & 22.6 & 19.6 & 81.3 & 84.3 & 53.9 & 21.7 & 79.8 & 42.9 & 24.2 & 46.8 & 46.0 & 53.3 \\
        CaCo~\cite{huang2021category} & 87.4 & 48.9 & 79.6 & 8.8 & 0.2 & 30.1 & 17.4 & 28.3 & 79.9 & 81.2 & 56.3 & 24.2 & 78.6 & 39.2 & 28.1 & 48.3 & 46.0 & 53.6 \\
        PixMatch~\cite{melas2021pixmatch} & 92.5 & 54.6 & 79.8 & 4.78 & 0.08 & 24.1 & 22.8 & 17.8 & 79.4 & 76.5 & 60.8 & 24.7 & 85.7 & 33.5 & 26.4 & 54.4 & 46.1 & 54.5 \\
        PLCA~\cite{kang2020pixel} & 82.6 & 29.0 & 81.0 & 11.2 & 0.2 & 33.6 & 24.9 & 18.3 & 82.8 & 82.3 & 62.1 & 26.5 & 85.6 & 48.9 & 26.8 & 52.2 & 46.8 & 54.0 \\
        DPL-Dual~\cite{Cheng_2021_ICCV} & 87.5 & 45.7 & 82.8 & 13.3 & 0.6 & 33.2 & 22.0 & 20.1 & 83.1 & 86.0 & 56.6 & 21.9 & 83.1 & 40.3 & 29.8 & 45.7 & 47.0 & 54.2 \\
        Seg-Uncert.~\cite{zheng_2021_IJCV} & 87.6 & 41.9 & 83.1 & 14.7 & 1.7 & 36.2 & 31.3 & 19.9 & 81.6 & 80.6 & 63.0 & 21.8 & 86.2 & 40.7 & 23.6 & 53.1 & 47.9 & 54.9  \\
        UPLR~\cite{Wang_2021_ICCV} & 79.4 & 34.6 & 83.5 & 19.3 & 2.8 & 35.3 & 32.1 & 26.9 & 78.8 & 79.6 & 66.6 & 30.3 & 86.1 & 36.6 & 19.5 & 56.9 & 48.0 & 54.6 \\
        CTF~\cite{Ma_2021_CVPR} & 75.7 & 30.0 & 81.9 & 11.5 & 2.5 & 35.3 & 18.0 & 32.7 & 86.2 & 90.1 & 65.1 & 33.2 & 83.3 & 36.5 & 35.3 & 54.3 & 48.2 & 55.5 \\
        DACS~\cite{DACS_2021_WACV} & 80.6 & 25.1 & 81.9 & 21.5 & 2.9 & 37.2 & 22.7 & 24.0 & 83.7 & \bf 90.8 & 67.6 & 38.3 & 82.9 & 38.9 & 28.5 & 47.6 & 48.3 & 54.8\\ 
        IAST~\cite{MeiZZZ20} & 81.9 & 41.5 & 83.3 & 17.7 & 4.6 & 32.3 & 30.9 & 28.8 & 83.4 & 85.0 & 65.5 & 30.8 & 86.5 & 38.2 & 33.1 & 52.7 & 49.8 & 57.0 \\
        KATPAN~\cite{Dong2021where} & 82.3 & 40.8 & 83.7 & 19.2 & 1.8 & 34.6 & 29.5 & 32.7 & 82.9 & 83.4 & 67.3 & 32.8 & 86.1 & 41.2 & 33.5 & 52.1 & 50.2 & 57.6 \\
        SAC~\cite{SAC_2021_CVPR} & 89.3 & 47.2 & 85.5 & 26.5 & 1.3 & 43.0 & 45.5 & 32.0 & 87.1 & 89.3 & 63.6 & 25.4 & 86.9 & 35.6 & 30.4 & 53.0 & 52.6 & 59.3 \\
        CorDA~\cite{CorDA_2021_ICCV} & \bf 93.3 & \bf 61.6 & 85.3 & 19.6 & \bf 5.1 & 37.8 & 36.6 & 42.8 & 84.9 & 90.4 & 69.7 & 41.8 & 85.6 & 38.4 & 32.6 & 53.9 & 55.0 & 62.8 \\
        ProDA~\cite{ProDA_2021_CVPR} & 87.8 & 45.7 & 84.6 & \bf 37.1 & 0.6 & \bf 44.0 & \bf 54.6 & 37.0 & \bf 88.1 & 84.4 & \bf 74.2 & 24.3 & 88.2 & 51.1 & 40.5 & 45.6 & 55.5 & 62.0 \\ 
        \hline
        {\bf \method (ProtoCL)} & 79.2 & 42.9 & \bf 85.6 & 9.9 & 4.2 & 38.0 & 52.5 & 53.3 & 80.6 & 81.2 & 73.7 & \bf 47.4 & 86.2 & 63.1 & 48.0 & 63.2 & 56.8 & 65.9 \\
        {\bf \method (BankCL)} & 76.7 & 34.3 & 84.9 & 18.7 & 2.9 & 38.4 & 51.8 & \bf 55.6 & 85.0 & 84.6 & 73.2 & 45.0 & \bf 89.7 & 63.7 & 50.5 & 63.8 & 57.4 & 66.1 \\
        {\bf \method (DistCL)} & 77.0 & 35.3 & 85.1 & 23.9 & 3.4 & 38.0 & 51.0 & 55.1 & 85.6 & 80.5 & 73.5 & 46.3 & 87.6 & \bf 69.7 & \bf 50.9 & \bf 66.5 & \bf 58.1 & \bf 66.5 \\
        \bottomrule[1.2pt]
          \end{tabular}} \VspaceAfter
\end{table*}
%##################################################################################################

%##################################################################################################
\begin{table*} \centering
  \caption{Comparison results of {\bf Cityscapes $\to$ Dark Zurich}. The DeepLab-V2 (D)~\cite{chen2018deeplab} and RefineNet (R)~\cite{lin2017refinenet} architecture with ResNet-101 trained on Cityscapes are used as Source Only baselines. The best result is highlighted in {\bf bold}.} \label{table:dark_zurich} \VspaceBefore
  \resizebox{\textwidth}{!}{
    \begin{tabular}{l|c|ccccccccccccccccccc|c}
      \toprule[1.2pt]
      Method & road & side. & buil. & wall & fence & pole & light & sign & veg. & terr. & sky & pers. & rider & car & truck & bus & train & mbike & bike & mIoU \\
      \hline
      Source Only & R & 68.8 & 23.2 & 46.8& 20.8 & 12.6 & 29.8 & 30.4 & 26.9 & 43.1 & 14.3 & 0.3 & 36.9 & 49.7 & 63.6 & 6.8 & 0.2 & 24.0 &33.6 &9.3 & 28.5\\
      DMAda~\cite{dai2018dark} & R & 75.5 & 29.1 & 48.6 & 21.3 & 14.3 & 34.3 & 36.8 & 29.9 & 49.4 & 13.8 & 0.4 & 43.3 & 50.2 & 69.4 & 18.4 & 0.0 & 27.6 & 34.9 & 11.9 & 32.1 \\
      GCMA~\cite{SakaridisDG19} & R & 81.7 & 46.9 & 58.8 & 22.0 & 20.0 & 41.2 & 40.5 & \bf 41.6 & 64.8 & 31.0 & 32.1 & \bf 53.5 & 47.5 & \bf 75.5 & 39.2 & 0.0 & 49.6 & 30.7 & 21.0 & 42.0 \\
      MGCDA~\cite{sakaridis2020mapguided} & R & 80.3 & 49.3 & 66.2 & 7.8 & 11.0 & 41.4 & 38.9 & 39.0 & 64.1 & 18.0 & 55.8 & 52.1 & 53.5 & 74.7 & \bf 66.0 & 0.0 & 37.5 & 29.1 & 22.7 & 42.5 \\
      DANNet~\cite{WU_2021_CVPR} & R & 90.0 & 54.0 & \bf 74.8 &\bf 41.0 & \bf 21.1 & 25.0 & 26.8 & 30.2 & \bf 72.0 & 26.2 & \bf 84.0 & 47.0 & 33.9 & 68.2 & 19.0 & \bf 0.3 & \bf 66.4 & 38.3 & 23.6 & 44.3 \\
      CDAda~\cite{xu2021cdada} & R & 90.5 & 60.6 & 67.9 & 37.0 & 19.3 & 42.9 & 36.4 & 35.3 & 66.9 & 24.4 & 79.8 & 45.4 & 42.9 & 70.8 & 51.7 & 0.0 & 29.7 & 27.7 & 26.2 & 45.0 \\
      \hline
      \hline 
      Source Only & D & 79.0 & 21.8 & 53.0 & 13.3 & 11.2 & 22.5 & 20.2 & 22.1 & 43.5 & 10.4 & 18.0 & 37.4 & 33.8 & 64.1 & 6.4 & 0.0 & 52.3 & 30.4 & 7.4 & 28.8\\
      AdaptSeg~\cite{tsai2018learning} & D & 86.1 & 44.2 & 55.1 & 22.2 & 4.8 & 21.1 & 5.6 & 16.7 & 37.2 & 8.4 & 1.2 & 35.9 & 26.7 & 68.2 & 45.1 & 0.0 & 50.1 & 33.9 & 15.6 & 30.4 \\
      AdvEnt~\cite{vu2019advent} & D & 85.8 & 37.9 & 55.5 & 27.7 & 14.5 & 23.1 & 14.0 & 21.1 & 32.1 & 8.7 & 2.0 & 39.9 & 16.6 & 64.0 & 13.8 & 0.0 & 58.8 & 28.5 & 20.7 & 29.7 \\
      BDL~\cite{li2019bidirectional} & D & 85.3 & 41.1 & 61.9 & 32.7 & 17.4 & 20.6 & 11.4 & 21.3 & 29.4 & 8.9 & 1.1 & 37.4 & 22.1 & 63.2 & 28.2 & 0.0 & 47.7 & \bf 39.4 & 15.7 & 30.8 \\
      DANNet~\cite{WU_2021_CVPR} & D & 88.6 & 53.4 & 69.8 & 34.0 & 20.0 & 25.0 & 31.5 & 35.9 & 69.5 & \bf 32.2 & 82.3 & 44.2 & 43.7 & 54.1 & 22.0 & 0.1 & 40.9 & 36.0 & 24.1 & 42.5 \\
      \hline
      \bf \method (ProtoCL) & D & 87.3 & 50.9 & 64.5 & 25.6 & 12.1 & 38.3 & 40.8 & 37.5 & 61.0 & 21.9 & 77.6 & 37.4 & 47.0 & 67.8 & 54.5 & 0.0 & 33.7 & 27.0 & 23.7 & 42.6 \\
      \bf \method (BankCL) & D & 88.5 & 54.8 & 66.5 & 25.1 & 13.5 & 40.0 & 39.6 & 40.8 & 62.5 & 25.1 & 79.0 & 37.8 & \bf 54.8 & 70.4 & 63.7 & 0.0 & 36.8 & 15.6 & 23.4 & 44.1 \\
      \bf \method (DistCL) & D & \bf 91.2 & \bf 61.3 & 67.0 & 28.5 & 15.5 & \bf 44.7 & \bf 44.3 & 41.3 & 65.4 & 22.5 & 80.4 & 41.3 & 52.4 & 71.2 & 39.3 & 0.0 & 39.6 & 27.5 & \bf 28.8 & \bf 45.4 \\
      \bottomrule[1.2pt]
    \end{tabular}} \VspaceAfter
\end{table*}
%##################################################################################################

\subsubsection{Comparisons with the state-of-the-arts}
\textbf{GTAV $\to$ Cityscapes.} We first present the adaptation results on the task of GTAV $\to$ Cityscapes in TABLE~\ref{table:gta}, with comparisons to the state-of-the-art DA approaches~\cite{ProDA_2021_CVPR,Charles2021confidence_estimation,CorDA_2021_ICCV,Ma_2021_CVPR,Dong2021where,huang2021category}, and the best results are highlighted in bold.
Overall, our \method (ProtoCL) sets the new state of the art. Particularly, we observe: (i) \method (DistCL) achieves 61.0\% mIoU, outperforming the baseline model trained merely on source data by a large margin of +22.4\% mIoU; (ii) Due to the rare presence of ``train" class in an image and its significant appearance difference across domains, our \method fails to predict them well. (iii) Adversarial training methods, e.g., AdaptSeg~\cite{tsai2018learning}, CLAN~\cite{luo2021category}, FADA~\cite{wang2020class}, can improve the transferability, but the effect is not as obvious as using self-training methods, e.g., Seg-Uncert.~\cite{zheng_2021_IJCV}, DACS~\cite{DACS_2021_WACV}, IAST~\cite{MeiZZZ20}, SAC~\cite{SAC_2021_CVPR}; (iv) On top of that, our \method (DistCL) beats the best-performing model, ProDA~\cite{ProDA_2021_CVPR}, by a considerable margin of +3.5\% mIoU, while ProDA has three complex training stages including warm up, self-training, and knowledge distillation.

Comparing the three variants of our framework, \method (ProtoCL) and \method (BankCL) also achieve remarkable mIoUs of 59.5\% and 60.4\% respectively. It is clear that BankCL and DistCL perform much better than ProtoCL, indicating features of higher quality are generated thanks to semantic concepts with greater diversity. It is worth reminding that methods built on memory banks are generally slower and demands more memory in training, while \method (DistCL) eases such burden and still manages to surpass \method (BankCL) at the same time.

\textbf{SYNTHIA $\to$ Cityscapes.}
As revealed in TABLE~\ref{table:synthetic}, our \method remains competitive on SYNTHIA $\to$ Cityscapes. \method (DistCL) attains 58.1\% mIoU and 66.5\% mIoU$^*$, achieving a significant gain of +24.6\% mIoU and +27.9\% mIoU$^*$ in comparison with ``Source Only'' model. It is noticeable that our \method (DistCL) ranks among the best in both mIoU and mIoU$^*$, outperforming ProDA~\cite{ProDA_2021_CVPR} by +2.6\% mIoU and CorDA~\cite{CorDA_2021_ICCV} by +3.7\% mIoU$^*$. The former is a multi-stage self-training framework and the latter combines auxiliary tasks, i.e., depth estimation, to facilitate knowledge transfer to the target domain. \method (ProtoCL/BankCL) also obtain a comparable performance in terms of mIoU$^*$ compared with \method (DistCL), but under-perform or tie with it in mIoU, indicating that a more class-balanced performance is done by \method (DistCL).

%##################################################################################################
\begin{figure}[t] \centering
    \includegraphics[width=0.46\textwidth]{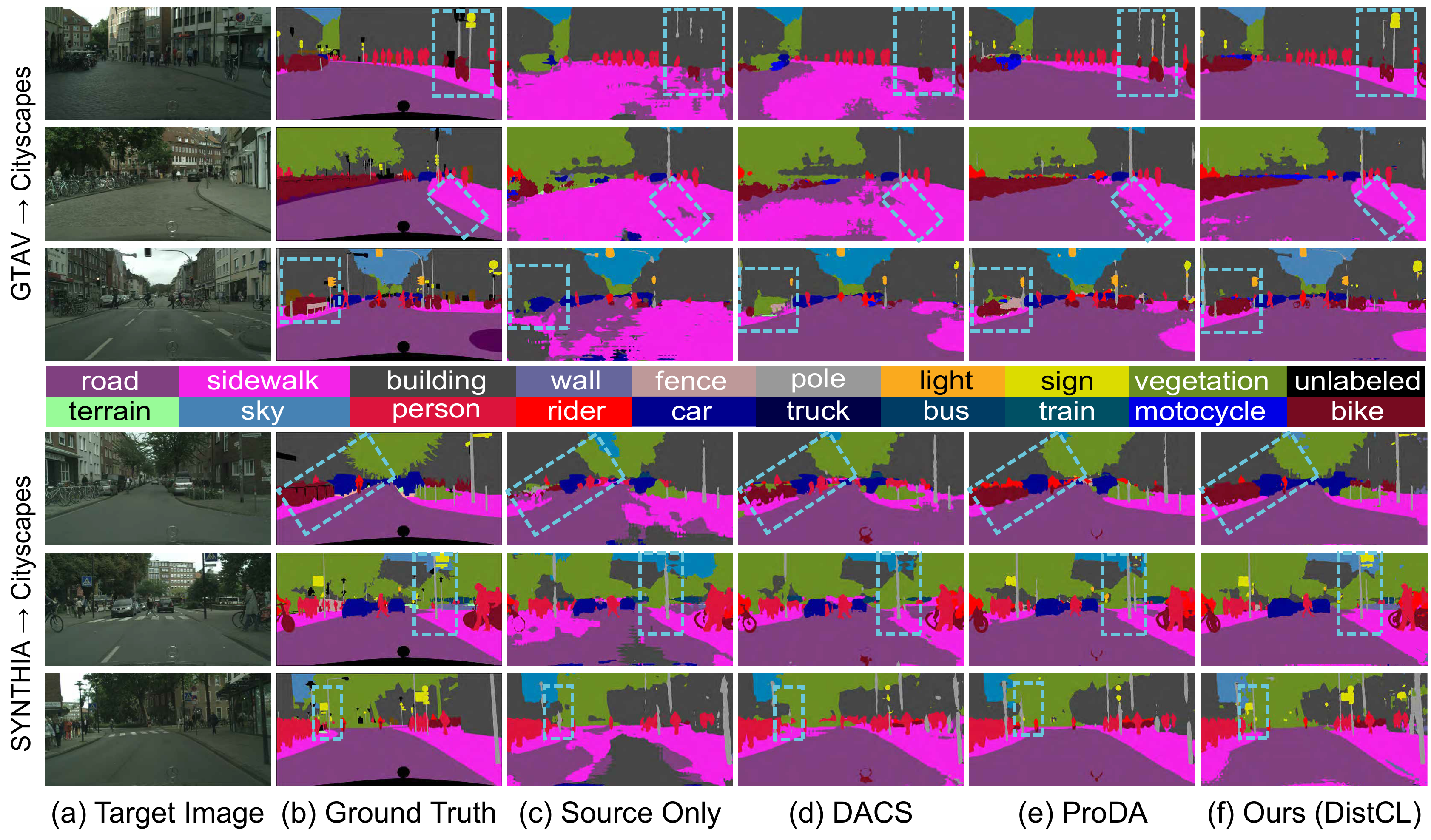} \VspaceBefore
    \caption{Qualitative results on Cityscapes (val). From left to right: target image, ground truth, the maps predicted by Source Only, DACS, ProDA and Ours (DistCL) are shown one by one. Our method shows a clear visual improvement.} \VspaceAfter
    \label{Fig_seg_map} 
\end{figure}
%##################################################################################################

%##################################################################################################
\begin{table*}[t] \centering
  \caption{Comparison results using Swin-B ViT~\cite{liu2021swin} and SegF. MiT-B5~\cite{xie2021segformer}. The best result is highlighted in {\bf bold}.} \VspaceBefore
  \label{table:transformer}
  \resizebox{\textwidth}{!}{
  \begin{tabular}{l|c c c c c c c c c c c c c c c c c c c|c}
  \toprule[1.2pt]
  \multicolumn{21}{c}{ (a) GTAV $\to$ Cityscapes}\\
  \hline 
  Method & road & side. & buil. & wall & fence & pole & light & sign & veg. & terr. & sky & pers. & rider & car & truck & bus & train & mbike & bike & mIoU \\
  \hline 
  Swin-B ViT (88M)                   & 63.3 & 28.6 & 68.3 & 16.8 & 23.4 & 37.8 & 51.0 & 34.3 & 83.8 & 42.1 & 85.7 & 68.5 & 25.4 & 83.5 & 36.3 & 17.7 & 2.9 & 36.1 & 42.3 & 44.6 \\
  TransDA-B~\cite{chen2022smoothing} & 94.7 & 64.2 & 89.2 & 48.1 & 45.8 & 50.1 & 60.2 & 40.8 & \bf 90.4 & 50.2 & \bf 93.7 & \bf 76.7 & \bf 47.6 & 92.5 & 56.8 & 60.1 & 47.6 & 49.6 & 55.4 & 63.9 \\
  \hline
  \hline 
  SegF. MiT-B5 (84.7M)            & 77.1 & 15.2 & 83.8 & 30.8 & 32.0 & 27.9 & 41.5 & 18.5 & 86.5 & 42.5 & 86.8 & 62.6 & 22.2 & 87.0 & 42.7 & 36.8 & 6.1 & 33.5 & 12.5 & 44.5 \\
  DAFormer~\cite{lukas2021daformer}  & 95.7 & 70.2 & 89.4 & 53.5 & 48.1 & 49.6 & 55.8 & 59.4 & 89.9 & 47.9 & 92.5 & 72.2 & 44.7 & 92.3 & 74.5 & 78.2 & \bf 65.1 & 55.9 & 61.8 & 68.3 \\
  {\bf \method (ProtoCL)}            & 96.1 & 72.9 & \bf 89.7 & 54.4 & 48.8 & 53.5 & 60.4 & \bf 65.3 & 90.0 & 48.4 & 91.6 & 75.2 & 47.1 & \bf 93.3 & 74.4 & 74.6 & 41.2 & 58.8 & \bf 65.9 & 68.5 \\
  {\bf \method (BankCL)}             & 96.3 & 73.6 & 89.6 & 53.7 & 47.8 & \bf 53.8 & \bf 60.8 & 60.0 & 89.9 & 48.8 & 91.5 & 74.6 & 45.1 & 93.1 & 74.8 & 73.8 & 51.5 & 60.3 & 65.3 & 68.7 \\
  {\bf \method (DistCL)}             & \bf 96.9 & \bf 76.7 & \bf 89.7 & \bf 55.5 & \bf 49.5 & 53.2 & 60.0 & 64.5 & 90.2 & \bf 50.3 & 90.8 & 74.5 & 44.2 & \bf 93.3 & \bf 77.0 & \bf 79.5 & 63.6 & \bf 61.0 & 65.3 & \bf 70.3 \\
  \hline
  \end{tabular}
  }
  
  \vfill 
  \resizebox{\textwidth}{!}{
  \begin{tabular}{l|c c c c c c c c c c c c c c c c|c c}
  \multicolumn{19}{c}{ (b) SYNTHIA $\to$ Cityscapes}\\
  \hline
  Method & road & side. & buil. & wall$^{*}$ & fence$^{*}$ & pole$^{*}$ & light & sign & veg. & sky & pers. & rider & car & bus & mbike & bike & mIoU & mIoU$^{*}$ \\
  \hline 
  Swin-B ViT (88M) & 57.3 & 33.8 & 56.0 & 6.3 & 0.2 & 33.8 & 35.5 & 18.9 & 79.9 & 74.8 & 63.1 & 10.9 & 78.3 & 39.0 & 20.8 & 19.4 & 39.2 & 45.2 \\
  TransDA-B~\cite{chen2022smoothing} & \bf 90.4 & \bf 54.8 & 86.4 & 31.1 & 1.7 & \bf 53.8 & \bf 61.1 & 37.1 & \bf 90.3 & \bf 93.0 & 71.2 & 25.3 & 92.3 & 66.0 & 44.4 & 49.8 & 59.3 & 66.3 \\
  \hline
  \hline 
  SegF. MiT-B5 (84.7M)            & 69.9 & 27.8 & 82.9 & 21.6 & 2.3 & 39.2 & 36.3 & 29.9 & 84.2 & 84.9 & 61.6 & 22.6 & 83.8 & 48.0 & 14.9 & 19.7 & 45.6 & 51.3 \\
  DAFormer~\cite{lukas2021daformer}  & 84.5 & 40.7 & 88.4 & \bf 41.5 & 6.5 & 50.0 & 55.0 & 54.6 & 86.0 & 89.8 & 73.2 & 48.2 & 87.2 & 53.2 & 53.9 & 61.7 & 60.9 & 67.4 \\
  {\bf \method (ProtoCL)}            & 85.9 & 45.5 & \bf 88.9 & 38.2 & 2.5 & 52.3 & 57.7 & 58.2 & 89.3 & 88.4 & 74.0 & 50.5 & 92.3 & 70.6 & \bf 56.2 & 56.7 & 62.9 & 70.3 \\
  {\bf \method (BankCL)}             & 88.2 & 49.3 & 88.6 & 36.1 & 4.7 & 53.1 & 58.9 & \bf 58.4 & 88.5 & 84.8 & 72.4 & 49.3 & \bf 92.8 & 76.3 & 55.5 & 55.2 & 63.3 & 70.6 \\
  {\bf \method (DistCL)}             & 87.0 & 52.6 & 88.5 & 40.6 & \bf 10.6 & 49.8 & 57.0 & 55.4 & 86.8 & 86.2 & \bf 75.4 & \bf 52.7 & 92.4 & \bf 78.9 & 53.0 & \bf 62.6 & \bf 64.3 & \bf 71.4 \\
  \hline
  \end{tabular}}
  
  \vfill
  \resizebox{\textwidth}{!}{
  \begin{threeparttable}
  \begin{tabular}{l|c c c c c c c c c c c c c c c c c c c|c}
  \multicolumn{21}{c}{ (c) Cityscapes $\to$ Dark Zurich}\\
  \hline
  Method & road & side. & buil. & wall & fence & pole & light & sign & veg. & terr. & sky & pers. & rider & car & truck & bus & train & mbike & bike & mIoU \\
  \hline 
  SegF. MiT-B5$^{\sharp}$ (84.7M)           & 80.3 & 37.1 & 57.5 & 28.1 & 7.9 & 35.5 & 33.2 & 29.3 & 41.7 & 14.8 & 4.7 & 48.9 & 48.0 & 66.6 & 5.7 & \bf 7.9 & 63.3 & 31.4 & 23.3 & 35.0 \\
  DAFormer$^{\sharp}$~\cite{lukas2021daformer}  & 92.0 & 63.0 & 67.2 & 28.9 & 13.1 & 44.0 & 42.0 & 42.3 & 70.7 & 28.2 & 83.6 & 51.1 & 39.1 & 76.4 & 31.7 & 0.0 & 78.3 & 43.9 & 26.5 & 48.5 \\
  {\bf \method (ProtoCL)}           & 90.1 & 57.7 & \bf 75.0 & \bf 34.9 & 16.4 & 53.5 & 47.0 & 47.8 & 70.1 & 31.7 & 84.1 & 57.3 & \bf 53.3 & 80.5 & 42.4 & 2.3 & 83.6 & 42.6 & \bf 30.1 & 52.7 \\
  {\bf \method (BankCL)}            & 91.1 & 61.2 & 73.4 & 31.9 & \bf 18.0 & 51.6 & 48.6 & 47.7 & 72.8 & \bf 33.0 & 85.5 & 57.0 & 51.1 & 80.6 & 48.4 & 3.1 & \bf 84.6 & \bf 45.3 & 28.2 & 53.3 \\
  {\bf \method (DistCL)}            & \bf 93.2 & \bf 68.1 & 73.7 & 32.8 & 16.3 & \bf 54.6 & \bf 49.5 & \bf 48.1 & \bf 74.2 & 31.0 & \bf 86.3 & \bf 57.9 & 50.9 & \bf 82.4 & \bf 52.2 & 1.3 & 83.8 & 43.9 & 29.8 & \bf 54.2 \\
  \bottomrule[1.2pt]
  \end{tabular}
  \begin{tablenotes}
      \item $^{\sharp}$ Implement according to source code.
    \end{tablenotes}
  \end{threeparttable}
  }
  \VspaceAfter
\end{table*}
%##################################################################################################

\textbf{Cityscapes $\to$ Dark Zurich.}
TABLE~\ref{table:dark_zurich} highlights the capability of our \method on the challenging daytime-to-nighttime task Cityscapes $\to$ Dark Zurich. To show that the current daytime-trained semantic segmentation models face significant performance degradation at night, we compare with AdaptSeg~\cite{tsai2018learning}, AdvEnt~\cite{vu2019advent}, and BDL~\cite{li2019bidirectional}, adopting DeepLab-V2 as backbone network. Our framework, especially \method (BankCL) and \method (DistCL), outperforms the comparison counterparts by a large margin.  The less powerful variant, \method (ProtoCL), is still able to win by a narrow margin when compared to the previous SOTA DANNet~\cite{WU_2021_CVPR}. Due to the huge domain divergence between daytime and nighttime scenarios, there are always two steps in prior methods. Take CDAda~\cite{xu2021cdada} as an example, it consists of inter-domain style transfer and intra-domain gradual self-training. While our \method aims at ensuring pixel-wise representation consistency between daytime and nighttime images, it is complementary to models designed for the nighttime and can still be trained in one stage. It is worth noting that our methods based on DeepLab-V2 are even superior or comparable to CDAda based on RefineNet~\cite{lin2017refinenet}, which further demonstrates the efficacy of our method.

%##################################################################################################
\begin{figure}[t] \centering
    \includegraphics[width=0.46\textwidth]{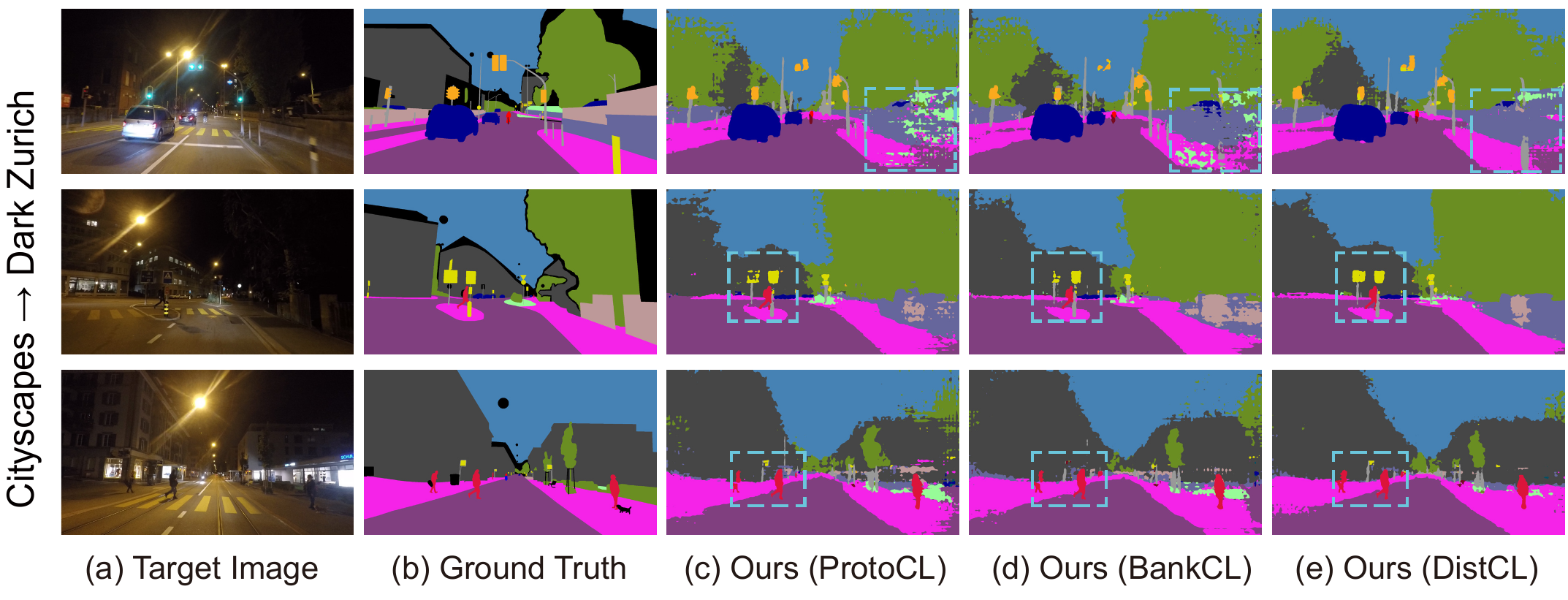}  \VspaceBefore
    \caption{Qualitative results on Dark Zurich (val). From left to right: target image, Ground Truth, the maps predicted by Ours (ProtoCL), Ours (BankCL) and Ours (DistCL).} \VspaceAfter
    \label{Fig_seg_map_dark} 
\end{figure}
%##################################################################################################

{\bf Qualitative results.}
In Fig.~\ref{Fig_seg_map}, we first visualize the segmentation results on two synthetic-to-real scenarios, predicted by our \method (DistCL), and compare our results to those predicted by the Source Only, DACS and ProDA models. The results predicted by \method (DistCL) are smoother and contain fewer spurious areas than those predicted by other models, showing that the performance has been largely improved. Next, as the daytime-to-nighttime task is far more challenging than the previous two, we further show several qualitative segmentation results in Fig.~\ref{Fig_seg_map_dark} to illustrate the advantage of \method (DistCL) over the other two variants \method (ProtoCL) and \method (BankCL).

\subsubsection{More Experimental Results}
{\bf Advanced network architecture.} Vision Transformer-based DA methods have been actively studied not long ago~\cite{chen2022smoothing,lukas2021daformer}. Hoyer et. al~\cite{lukas2021daformer} analyze different architectures for adaptation and propose a new architecture, DAFormer, based on a Transformer encoder~\cite{xie2021segformer} and a context-aware fusion decoder. Lately, Chen et al.~\cite{chen2022smoothing} introduce a momentum network and dynamic of discrepancy measurement to smooth the learning dynamics for target data. Therefore, we further adopt one of architectures such as DAFormer~\cite{lukas2021daformer}, to support our claims. Inspired by a multi-level context-aware feature fusion decoder, we also fuse all stacked multi-level features from the decoder to provide valuable concepts for contrastive learning. From TABLE~\ref{table:transformer}, we have the following observations: (i) Approaches based on Transformer perform generally better than those based on DeepLab-V2, confirming the strength of these advanced architectures; (ii) Our \method is still competitive on the new architecture. All variants of \method achieve an extraordinary improvement of around +20\% mIoU on each task when compared with the models trained merely on source data, i.e., Swin-B ViT~\cite{liu2021swin} and SegF. MiT-B5~\cite{xie2021segformer}; (iii) \method (DistCL) improves the state-of-the-art DAFormer by +2.0\% mIoU for GTAV $\to$ Cityscapes, +3.4\% mIoU for SYNTHIA $\to$ Cityscapes, and +5.7\% mIoU for Cityscapes $\to$ Dark Zurich.

%##################################################################################################
\begin{table} \centering
  \caption{Comparison results of Cityscapes $\rightarrow$ Dark Zurich trained models for generalization on two unseen target domains: Nighttime Driving and BDD100k-night test sets.}\VspaceBefore
  \label{tab:night_generalization}
  \resizebox{\linewidth}{!}{
  \begin{tabular}{lcccc|c}
    \toprule[1.2pt]
    Method && Dark Zurich & Nighttime Driving & BDD100k-night & Cityscapes \\
    \hline
    DMAda (RefineNet)~\cite{dai2018dark} && 32.1 & 36.1 & 28.3 & - \\
    GCMA (RefineNet)~\cite{SakaridisDG19} && 42.0 & 45.6 & 33.2 & - \\
    MGCDA (RefineNet)~\cite{sakaridis2020mapguided} && 42.5 & 49.4 & 34.9 & -\\
    CDAda (RefineNet)~\cite{xu2021cdada} && 45.0 & 50.9 & 33.8 & - \\
    \hline
    \hline
    SegF. MiT-B5~\cite{xie2021segformer} && 35.0 & 46.9 & 34.0 &  76.8 \\
    DAFormer~\cite{lukas2021daformer} && 48.5 & 51.8 & 33.9 & 76.4 \\
    \bf\method (ProtoCL) && 52.7 & 55.5 & 37.5 & \bf 79.5 \\
    \bf\method (BankCL) && 53.3 & 54.9 & 39.1 & 78.9 \\
    \bf\method (DistCL) && \bf 54.2 & \textbf{56.9} & \textbf{40.6} & 78.9 \\
    \bottomrule[1.2pt]
  \end{tabular}} \VspaceAfter
\end{table}
%##################################################################################################

%##################################################################################################
\begin{table}[t] \centering
  \caption{Experiments over weather DA object detection: Cityscapes $\rightarrow$ Foggy Cityscapes based on Faster R-CNN.} \VspaceBefore
  \label{table:detection}
  \resizebox{\linewidth}{!}{
  \begin{tabular}{lccccccccc}
    \toprule[1.2pt]
    Method & person & rider & car & truck & bus & train & mcycle & bicycle & mAP${^r_{0.5}}$ \\
    \hline
    Faster R-CNN~\cite{ren2015faster} & 17.8 & 23.6 & 27.1 & 11.9 & 23.8 & 9.1 & 14.4 & 22.8 & 18.8 \\
    DA-Faster~\cite{chen2018domain} &25.0 &31.0 &40.5 &22.1 &35.3 &20.2 &20.0 &27.1 &27.6\\ 
    SW-Faster~\cite{saito2019strong} & 36.2 &35.3 &43.5 &30.0 &29.9 &42.3 & 32.6 & 24.5 & 34.3\\
    PDA~\cite{HsuYTHT0020} & 36.0 & 45.5 & 54.4 & 24.3 & 44.1 & 25.8 & 29.1 & 35.9 & 36.9\\
    EveryPixelMatters~\cite{HsuTLY20} & 41.9 & 38.7 & 56.7 & 22.6 & 41.5 & 26.8 & 24.6 & 35.5 & 36.0 \\
    \hline
    \bf  \method (ProtoCL) & 41.9 & 40.4 & 58.2 & 24.9 & 40.3 & 32.6 & 25.5 & 35.0 & 37.4 \\
    \bf \method (BankCL) & 43.8 & 42.8 & 57.9 & 22.5 & 43.0 & 26.4 & 27.0 & 38.9 & 37.8 \\
    \bf \method (DistCL) & 43.9 & 41.2 & 57.5 & 25.1 & 42.8 & 26.1 & 29.1 & 39.1 & \bf 38.1 \\
    \hline
    Faster RCNN (oracle) & 47.4 & 40.8 & 66.8 & 27.2 & 48.2 & 32.4 & 31.2 & 38.3 & 41.5 \\
    \bottomrule[1.2pt]
  \end{tabular} } \VspaceAfter
\end{table}
%##################################################################################################

{\bf Generalization to unseen domains.} In TABLE~\ref{table:dark_zurich} and TABLE~\ref{table:transformer}(c), we have benchmarked our method on the Dark Zurich test. To showcase the better generalization of \method, the trained Dark Zurich models are also tested on two unseen target domains, i.e., Nighttime Driving~\cite{dai2018dark} and BDD100k-night~\cite{yu2020bdd100k}. From TABLE~\ref{tab:night_generalization}, we can find that the generalization ability of previous self-training methods is limited, and they often fail to transfer well to unseen domains or concepts. On the contrary, our \method markedly improves over the sophisticated baselines. Notably, \method (DistCL) achieves mIoUs of 56.9\% and 40.6\%, respectively, releasing the newest records on both. Another interesting finding is that our SePiCo indeed boosts the segmentation performance on the source domain even compared with SegF. MiT-B5 (Source Only) model that is only trained on source images. There is some evidence that a well-structured pixel embedding space provides the best of both worlds: reducing distribution shift, plus promoting the source task.

{\bf Adaptation for object detection.} We further extend our \method to a weather adaptive object detection task, i.e., Cityscapes~\cite{Cordts2016Cityscapes} $\to$ Foggy Cityscapes~\cite{SakaridisDHG18}. More specifically, Foggy Cityscapes is a synthetic foggy dataset that applies simulated fog to scenes of Cityscapes. Build upon~\cite{HsuTLY20}, we employ Faster R-CNN~\cite{ren2015faster} with VGG-16~\cite{simonyan2015vgg} as the backbone. The model is trained with learning rate of $5\times 10^{-3}$, momentum of 0.9 and weight decay of $5\times 10^{-4}$. The image's shorter side is set to 800 and RoIAlign is employed for feature extraction. From TABLE~\ref{table:detection}, we observe that \method achieves comparable results with other task-specific and well-optimized detection algorithms~\cite{HsuTLY20,saito2019strong,HsuYTHT0020}.

\subsection{Ablation Studies}
\label{sec:ablation}
\begin{figure}[t]
    \centering
    \subfloat[Study on N\_crop.]{
      \begin{minipage}[b]{0.44\linewidth} \centering  
        \includegraphics[width=\linewidth]{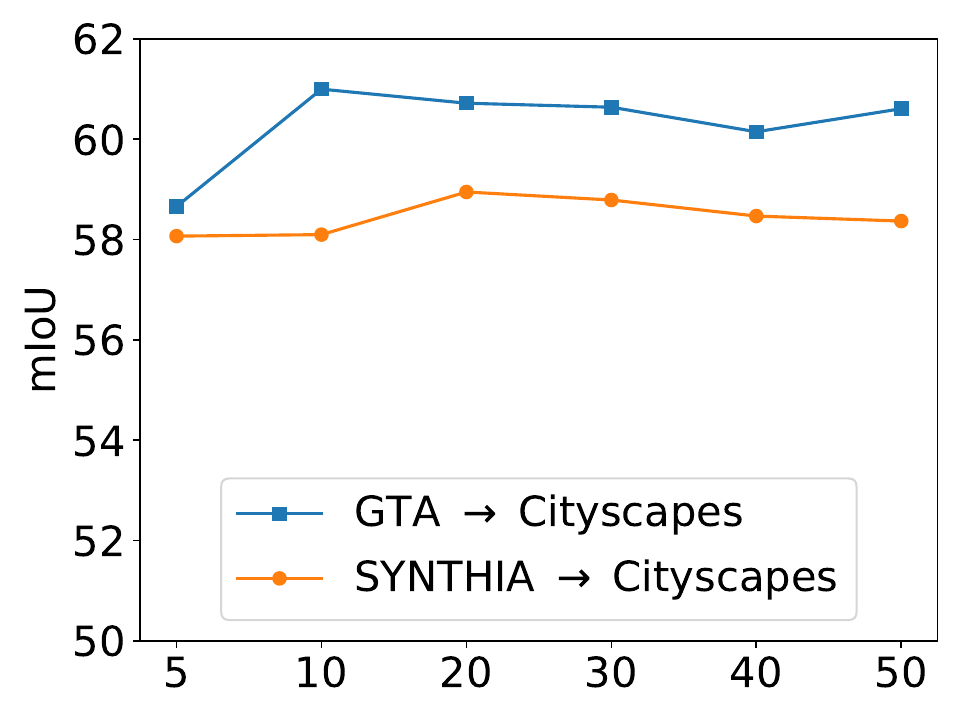} 
      \end{minipage}
    }
    \subfloat[Study on cat\_max\_ratio.]{
      \begin{minipage}[b]{0.44\linewidth} \centering  
        \includegraphics[width=\linewidth]{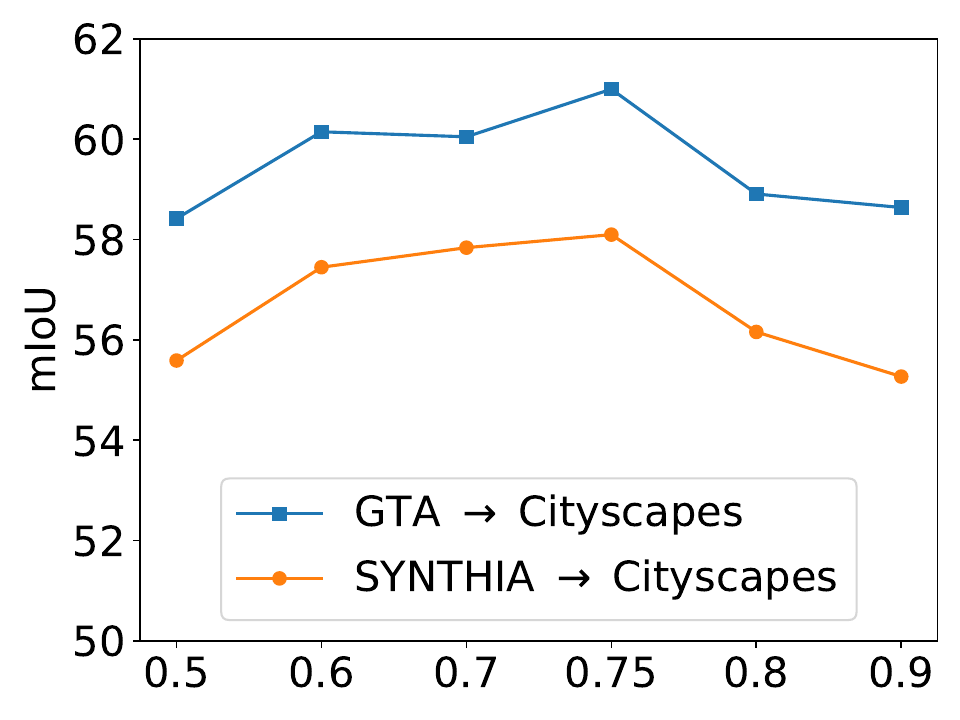} 
      \end{minipage}%
    } \VspaceBefore
    \caption{Parameter sensitivity analysis for CBC.} \VspaceAfter
    \label{Fig_cbc_param}
\end{figure}

We evaluate the contribution of each component present in our one-stage framework. Specifically, we testify \method on the task of GTAV $\to$ Cityscapes, and the results are reported in TABLE~\ref{table:ablation}, TABLE~\ref{table:teacher_bank}, TABLE~\ref{table:multi_feature} and Fig.~\ref{Fig_cbc_param}. As can be seen, each of these components contributes to the ultimate success. Eventually, we achieve 49.8\% and 61.0\% mIoU under ``w/o self-training'' and ``\method (DistCL)'' respectively, outperforming the corresponding baselines by +11.2\% and +8.9\%.

{\bf Effect of semantic-guided pixel contrast.} As discussed in Section~\ref{sec:contrastive_adaptation}, centroid-aware and distribution-aware pixel contrast can build up stronger intra-/inter-category connections and minimize the domain divergence efficiently. We validate the performance increments by separately training models with and without self-training. As shown in Table~\ref{table:ablation}, contrastive learning alone can improve the segmentation performance, but the effect is not as noticeable as using self-training (48.5\% mIoU vs. 52.1\% mIoU). When they are adopted properly in a unified pipeline, the full potential of the model is released, further promoting gains of +7.2\% mIoU. The results imply the effect and necessity of representation learning for the classical self-training paradigm.

%##################################################################################################
\begin{table}[t] \centering
    \caption{Ablation study on GTAV $\to$ Cityscapes. All models are \textbf{trained end-to-end} in a total of 40k iterations.} \VspaceBefore
    \label{table:ablation}
    \resizebox{0.46\textwidth}{!}{
        \begin{tabular}{l|cccc|cc}
            \toprule[1.2pt]
            Method & \loss{\ssl} & \loss{\cl} & \loss{\reg} & CBC & mIoU & $\Delta$ \\
            \hline
            \multirow{4}{*}{w/o self-training} & & &  & & 38.6$_{\pm0.5}$ & - \\
            & & \checkmark & & & 48.5$_{\pm0.8}$ & 9.9 \\
            & & \checkmark & \checkmark & & 49.2$_{\pm0.7}$ & 10.6 \\
            & & \checkmark & \checkmark & \checkmark & 49.8$_{\pm0.4}$ & 11.2 \\
            \hline
            \multirow{4}{*}{\method (DistCL)} & \checkmark & & & & 52.1$_{\pm2.0}$ & -\\
            & \checkmark & \checkmark &  &  & 59.3$_{\pm1.7}$ & 7.2 \\
            & \checkmark & \checkmark & \checkmark &  & 60.4$_{\pm1.3}$ & 8.3 \\
            & \checkmark & \checkmark & \checkmark & \checkmark & \bf 61.0$_{\pm0.7}$ & 8.9 \\
            \bottomrule[1.2pt]
        \end{tabular}} \VspaceAfter
\end{table}
%##################################################################################################

%##################################################################################################
%##################################################################################################

%##################################################################################################
\begin{figure}[t]
    \centering
    \includegraphics[width=0.48\textwidth]{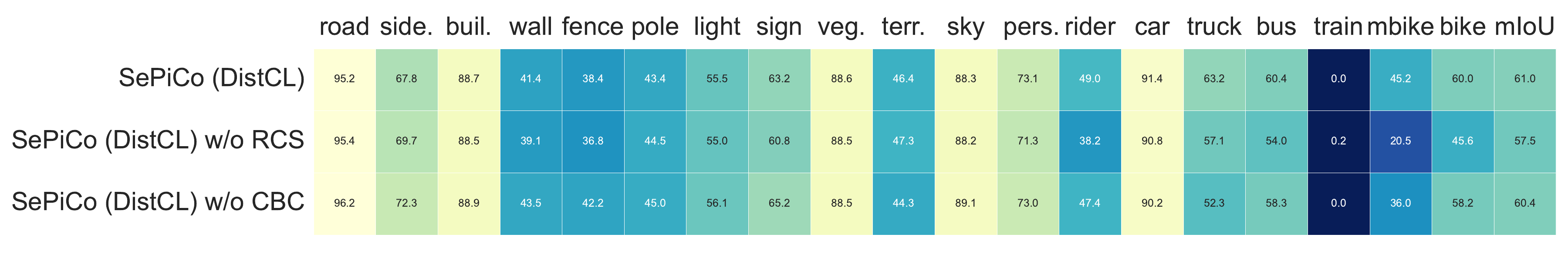}
    \VspaceBefore
    \caption{Comparison results of per-class IoU for CBC and RCS.}
    \label{Fig_class_heatmap}
    \VspaceAfter
\end{figure}
%##################################################################################################

%##################################################################################################
\begin{table}[!htbp] \centering
    \caption{Effect of the teacher network and the bank size  on GTAV $\to$ Cityscapes. The default choice is colored in \colorbox{Gray}{gray}.}  \label{table:teacher_bank}
    \VspaceBefore
    \subfloat[Effect of the teacher network.]{
    \begin{minipage}{0.42\textwidth} \label{table:param_teacher}
        \centering
        \resizebox{\textwidth}{!}{
            \begin{tabular}{ccccc|c}
                \toprule[1.2pt]
                 & \multicolumn{4}{c|}{w/ teacher} &  w/o teacher \\
                $\beta$ & 0.99 & \cellcolor{Gray}0.999 & 0.9995 & 0.9999 & 0.0 \\ 
                \hline
                mIoU & 60.8 & \bf 61.0 & 60.6 & 60.7 & 56.1 \\
                \bottomrule
            \end{tabular}
        }
    \end{minipage}
    }
    \hfill
    \subfloat[Effect of the bank size $B$.]{
    \begin{minipage}{0.42\textwidth} \label{table:param_bank}
        \resizebox{\textwidth}{!}{
            \begin{tabular}{ccccc|c}
                \toprule[1.2pt]
                & \multicolumn{4}{c|}{\method (BankCL)} & \method (DistCL) \\
                $B$ & 50 & 100 & \cellcolor{Gray}{200} & 500 & $\infty$\\ 
                \hline
                mIoU & 59.4 & 59.5 & \bf 59.8 & 59.7 & 61.0 \\
                \bottomrule
            \end{tabular}
        }
    \end{minipage}  
    }
    \VspaceAfter
\end{table}
%##################################################################################################

%##################################################################################################
\begin{table}[t]
    \centering
    \caption{Ablation of feature selection in \method variants for GTAV $\to$ Cityscapes based on DeepLab-V2.}
    \VspaceBefore
    \label{table:multi_feature}
    \resizebox{0.44\textwidth}{!}{
        \begin{tabular}{l|cccc|c}
            \toprule[1.2pt]
            Method & layer 1 & layer 2 & layer 3 & layer 4 & mIoU \\
            \hline
            \multirow{5}{*}{\bf \method (ProtoCL)}  & \checkmark & &  & & 58.7 \\ 
            &  & \checkmark &  &  & 58.6  \\ 
            &  &  & \checkmark &  & 58.6 \\  
            &  &  & & \cellcolor{Gray}\checkmark & \bf 58.8 \\  
            & \multicolumn{4}{c|}{\{1,2,3,4\}-fusion} & 58.4  \\ 
            \hline
            \multirow{5}{*}{\bf \method (BankCL)} & \checkmark & &  & & 58.5 \\ 
            && \checkmark &  &  & 58.4  \\ 
            &&  & \checkmark &  & 58.0  \\ 
            &&  &  & \cellcolor{Gray}\checkmark & \bf 59.8  \\ 
            & \multicolumn{4}{c|}{\{1,2,3,4\}-fusion} & 58.6 \\ 
            \hline
            \multirow{5}{*}{\bf \method (DistCL)} & \checkmark & & & & 60.7 \\
            & & \checkmark &  &  & 59.3 \\
            & &  & \checkmark &  & 58.8 \\
            & &  &  & \cellcolor{Gray}\checkmark & \bf 61.0 \\
            & \multicolumn{4}{c|}{\{1,2,3,4\}-fusion} & 60.2 \\
            \bottomrule
        \end{tabular}
    }
 \VspaceAfter
\end{table}
%##################################################################################################

%##################################################################################################
\begin{figure*}[t]
    \centering
    \includegraphics[width=0.92\textwidth]{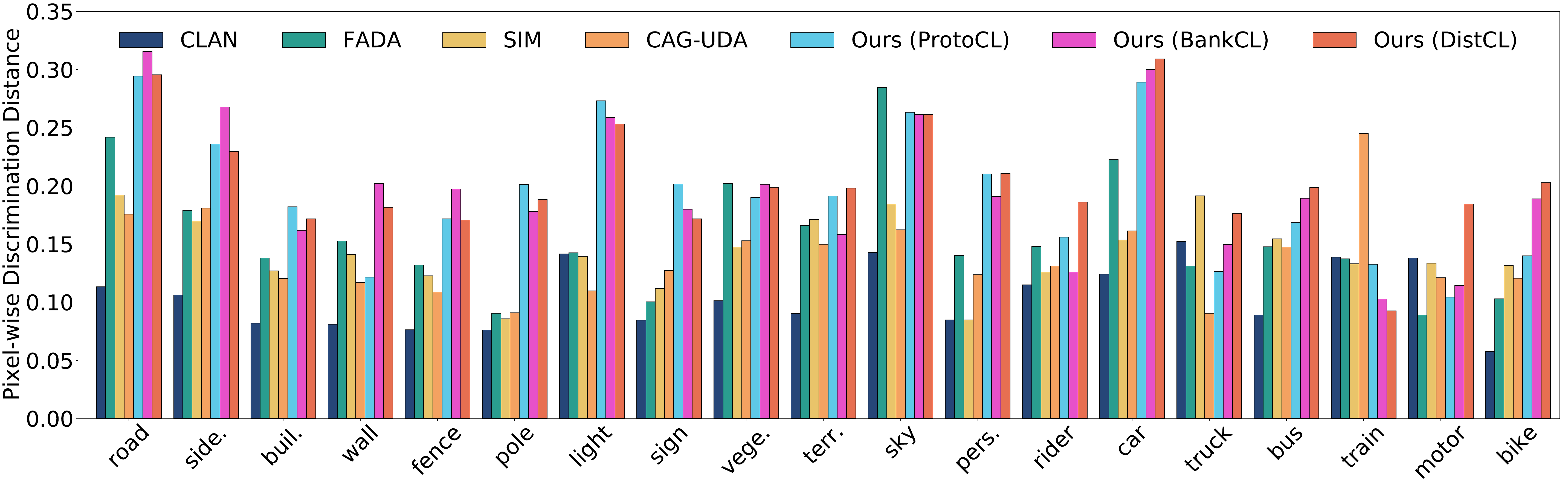}
    \VspaceBefore
    \caption{Quantitative analysis of the discrimination of features. For each class, we show the values of pixel-wise discrimination distance (PDD) as defined in Eq.~\eqref{eq:pdd} on Cityscapes validation set. These comparison results are from 1) category adversarial learning methods, i.e, CLAN and FADA; 2) category centroid-based alignment methods i.e., SIM and CAG-UDA; 3) pixel contrast methods, i.e., Ours (ProtoCL/BankCL/DistCL), respectively. A high PDD suggests the pixel-wise representations of same category are clustered densely while the distance between different categories is relatively large.}
    \label{Fig_pdd}
    \VspaceAfter
\end{figure*}
%##################################################################################################

\textbf{Effect of \loss{\reg}.} We study the advantages of diversity-promoting regularization term \loss{\reg} in TABLE~\ref{table:ablation}.  It is clearly shown that using \loss{\reg} also brings an extra increase (+0.7\% mIoU and +1.1\% mIoU, respectively), verifying the effectiveness of smoothing the learned representations.

{\bf Effect of class-balanced cropping (CBC).} We first remove the class-balanced cropping discussed in Section~\ref{sec:class_balance_cropping} to verify its necessity. As shown in TABLE~\ref{table:ablation}, as expected, the mIoU of the adapted model decreases moderately without CBC, supporting the importance of class imbalance cropping. Furthermore, we test CBC from parameter sensitivity and compare it with other strategies. From Fig.~\ref{Fig_cbc_param}, we can observe that increasing cropping times (N\_crop) helps us to select more class-balanced regions and bring slight gains. However, to increase the training efficiency, we just make 10 croppings for all experiments. As for cat\_max\_ratio, a high threshold (e.g., $>$ 0.9) will result in too many futile candidates and, conversely, a low threshold (e.g., $<$ 0.5) will result in no candidate (In this case, the first crop will be selected). Both cases invalidate CBC, which leads to random cropping. Thus, cat\_max\_ratio is default set to 0.75.

Furthermore, Fig.~\ref{Fig_class_heatmap} shows the results of class-wise IoU of ablating two class-balanced strategies (RCS and CBC), respectively. As seen, RCS mainly improves the performance of minority classes (e.g., large gains on ``rider", ``mbike" and ``bike"). While our CBC mainly aims to balance the IoUs among all categories, in which some categories such as ``side", ``wall", and ``fence" even show slight decreases in IoU while others show tremendous increases in IoU. In summary, the RCS is a direct yes effective class-balanced strategy since supervised annotations are utilized. In contrast, the proposed CBC serves as an alternative choice in the absence of annotations but is also able to balance the IoUs among all categories and improve the overall performance.

\textbf{Effect of the teacher network.} The teacher-student architecture is frequently adopted to introduce a strong regularization during training~\cite{TarvainenV17}. Essentially, a larger momentum value $\beta$ indicates a stronger effect from the teacher net. We adjust $\beta$ to change the amount of regularization and report the results in TABLE~\ref{table:param_teacher}. A performance gain of more than +3.0\% mIoU is brought about by the teacher net, confirming its efficacy. Thus $\beta$ is fixed to 0.999 for proper regularization.

\textbf{Effect of the bank size $B$.} TABLE~\ref{table:param_bank} lists the effect of bank size for \method (BankCL). As we enlarge the memory bank from 50 to 500, a gradual gain can be witnessed in performance, with a slight drop when $B$=500. Generally, a larger bank size means more diversity in semantic concepts, leading to better performance. However, a huge bank will result in prolonged retention of outdated representations, which may exert a negative effect on pixel-level guidance. In comparison, \method (DistCL) overcomes this issue by using the distribution to simulate infinite bank size on-the-fly, thus exceeding \method (BankCL) by a considerable margin.

\textbf{Effect of multi-level features.} We provide the performance of applying \method to intermediate layers. In particular, there are four residual blocks in the original ResNet-101 backbone~\cite{he2016deep}. The four layers (denoted by layer 1 - layer 4) are taken from the output of each residual block and \{1,2,3,4\}-fusion means that features from all four layers are concatenated together. Interestingly, features from layer 1 also exhibit distinctive information. This is expected because earlier features provide valuable low-level concepts for semantic segmentation at a high resolution. However, if we fuse all the features for adaptation, the results are slightly degraded, which is different from the Transformer-based architecture. We conjecture that the ViTs have more similarity between the representations obtained in shallow and deep layers compared to CNNs.
Overall, the features from layer 4 prove to be the best choice for all three variants of \method. It can be seen that \method (DistCL) performs nearly equally well while adopting features from the last layer or the fusion of multiple layers, indicating that the distribution indeed increases the diversity of features and is more robust.

%##################################################################################################
\begin{table*}
  \centering
  \subfloat[Study on \hparam{\cl}.]{
  \begin{minipage}{0.3\textwidth}
      \label{table:param_cl}
      \centering
      \resizebox{\textwidth}{!}{
          \begin{tabular}{c|ccccc}
              \hparam{\cl} & 0.01 & 0.1 & 0.5 & \cellcolor{Gray}1.0 & 2.0 \\
              \hline
              G $\rightarrow$ C & 58.9 & 59.9 & 60.9 & \bf 61.0 & 59.3 \\
              S $\rightarrow$ C & 57.3 & 57.4 & 57.8 & \bf 58.1 & 57.9 \\
          \end{tabular}
      }
  \end{minipage}
  }
  \subfloat[Study on \hparam{\reg}.]{
  \begin{minipage}{0.3\textwidth}
      \label{table:param_reg}
      \centering
      \resizebox{\textwidth}{!}{
          \begin{tabular}{c|ccccc}
              \hparam{\reg} & 0.01 & 0.1 & 0.5 & \cellcolor{Gray}1.0 & 2.0 \\
              \hline
           G  $\rightarrow$ C & 59.5 & 59.3 & 59.3 & \bf 61.0 & 58.7 \\
          S $\rightarrow$ C & 57.8 & 57.9 & \bf58.2 & 58.1 & 57.6 \\
          \end{tabular}
      }
  \end{minipage}
  }
  \subfloat[Study on $L_w$.]{
  \begin{minipage}{0.3\textwidth}
      \label{table:param_start_iter}
      \centering
      \resizebox{\textwidth}{!}{
          \begin{tabular}{c|ccccc}
              $L_w$ & 0 & 1500 & \cellcolor{Gray}3000 & 5000 & 10000 \\
              \hline
              G $\rightarrow$ C & 59.2 & 59.5 & \bf 61.0 & 59.7 & 58.4 \\
              S $\rightarrow$ C & 57.2 & 58.0 & \bf58.1 & 57.8 & 57.3 \\
          \end{tabular}
      }
  \end{minipage}
  }
  \VspaceBefore
  \caption{Parameter sensitivity on GTAV $\to$ Cityscapes (G $\to$ C) and SYNTHIA $\to$ Cityscapes (S $\to$ C) tasks.}
  \VspaceAfter
\end{table*}
%##################################################################################################

\subsection{Further Analysis}
\label{sec:analysis}
\subsubsection{Pixel-wise Discrimination Distance}
To verify whether our adaptation framework can yield a discriminative embedding space, we design a metric to take a closer look at what degree the pixel-wise representations are aligned. In the literature, CLAN~\cite{luo2021category} defines a Cluster Center Distance as the ratio of the intra-category distance between the initial model and the aligned model and FADA~\cite{wang2020class} proposes a new Class Center Distance to consider inter-category distance. To better evaluate the effectiveness of pixel-wise representation alignment, we introduce a new Pixel-wise Discrimination Distance (PDD) by taking intra- and inter-category affinities of pixel representations into account. Formally, a PDD value for category $k$ is given by:
\begin{small}
\begin{align}
    PDD(k) = \frac{1}{|\Lambda^k|} \sum_{x\in \Lambda^k} \frac{sim(x, \mu^k)}{\sum_{i=1,i\neq k}^K sim(x, \mu^{i})} \,,
    \label{eq:pdd}
\end{align}
\end{small}%
where $sim(\cdot, \cdot)$ is the similarity metric, and we adopt cosine similarity. $\Lambda^k$ denotes the pixel set that contains all the pixel representations belonging to the $k^{th}$ semantic class.

\begin{figure}[t] \centering
    \includegraphics[width=0.42\textwidth]{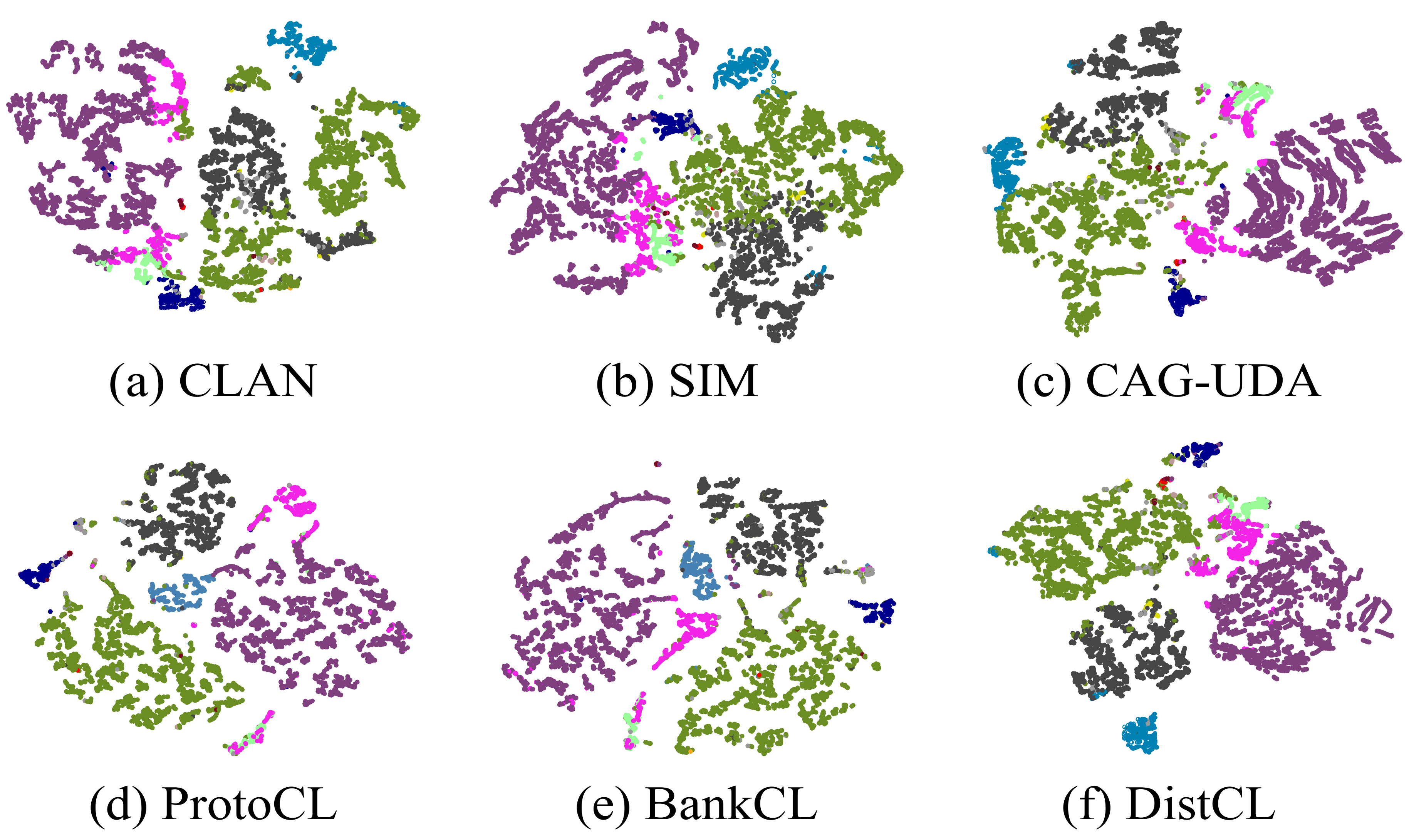} \VspaceBefore
    \caption{t-SNE analysis of existing comparable alignment methods and our \method. As seen, the proposed pixel contrast objectives (ProtoCL/BankCL/DistCL) beget well-structured embedding spaces. Please zoom in for details.} \VspaceAfter
    \label{Fig_tsne} 
\end{figure}

With PDD, we could investigate the relative magnitude of inter-category and intra-category pixel feature distances. Specifically, we calculate the PDD on the whole Cityscapes validate set and compare PDD values with other state-of-the-art category alignment methods: CLAN~\cite{luo2021category} and FADA~\cite{wang2020class} for category-level adversarial training and SIM~\cite{wang2020differential} and CAG-UDA~\cite{zhang2019category} for category centroid based counterparts that do not tackle the distance between different category features. As shown in Fig.~\ref{Fig_pdd}, we observe that: (i) CLAN and FADA could not cope well with the distance between different category features, thus obtaining lower PDD values. (ii) Both SIM and CAG-UDA adopt category anchors computed on the source domain to guide the alignment but they do not regularize the distance among different category features. Thus, the PDD values of some categories such as ``road", ``wall", ``light" and ``car" are even lower than those of adversarial training methods while the PDD values of some categories are sometimes higher. (iii) Considering cross-domain pixel contrast, our SePiCo (ProtoCL/BankCL/DistCL) can achieve much higher PDD values in most categories. Based on these quantitative results, together with the t-SNE analysis in Fig.~\ref{Fig_tsne}, it is clear that our \method can achieve better pixel-wise category alignment and largely improve the pixel-wise accuracy of predictions.

%##################################################################################################
%##################################################################################################

\subsubsection{t-SNE Visualization}
To better develop intuition, we draw t-SNE visualizations~\cite{maaten2008visualizing} of learned representations for three competitive category alignment methods (CLAN~\cite{luo2021category}, SIM~\cite{wang2020differential}, CAG-UDA~\cite{zhang2019category}) and compare them with all variants of our \method (ProtoCL, BankCL, DistCL) in Fig.~\ref{Fig_tsne}. With this in mind, we first randomly select an image from target domain and then map its high-dimensional latent feature representations to a 2D space. From the t-SNE visualizations, we can observe that (i) Existing category alignment methods could produce separated features, but it may be hard for dense prediction since the margins between different category features are not obvious and the distribution is still dispersed; (ii) When we apply pixel contrast, features among different categories are better separated, demonstrating that the semantic distributions can provide correct supervision signal for target data; (iii) More importantly, the representations of \method (DistCL) exhibit clear clusters, revealing the discriminative capability of the distribution-aware contrastive adaptation.

\subsubsection{Parameter Sensitivity}
\label{sec:parameter_sensitivity}
We conduct parameter sensitivity analysis to evaluate the sensitivity of \method (DistCL) on two synthetic-to-real benchmarks. As shown in TABLE~\ref{table:param_cl}, \ref{table:param_reg} and \ref{table:param_start_iter}, we select loss weights \hparam{\cl} and \hparam{\reg} $\in \{0.01\,, 0.1\,, 0.5\,, 1.0\,, 2.0\}$, the iteration at which to start contrastive learning $L_w \in \{0\,, 1500\,, 3000\,, 5000\,, 10000\}$, respectively. While altering \hparam{\cl} and \hparam{\reg} in a large range, we find that both losses are slightly sensitive to their assigned weight on GTAV $\to$ Cityscapes, and probably their relative weight due to their resemblance. Nevertheless, our method keeps outperforming the previous SOTA in different compositions of loss weights. We also explore the sensitivity of our method on the iteration to start contrastive learning and observe that \method (DistCL) is relatively robust to $L_w$, peaking at $L_w=3000$. The result could be attributed to better category information learned through warm-up iterations.

%##################################################################################################
\begin{figure}[t]
    \centering
    \includegraphics[width=0.4\textwidth]{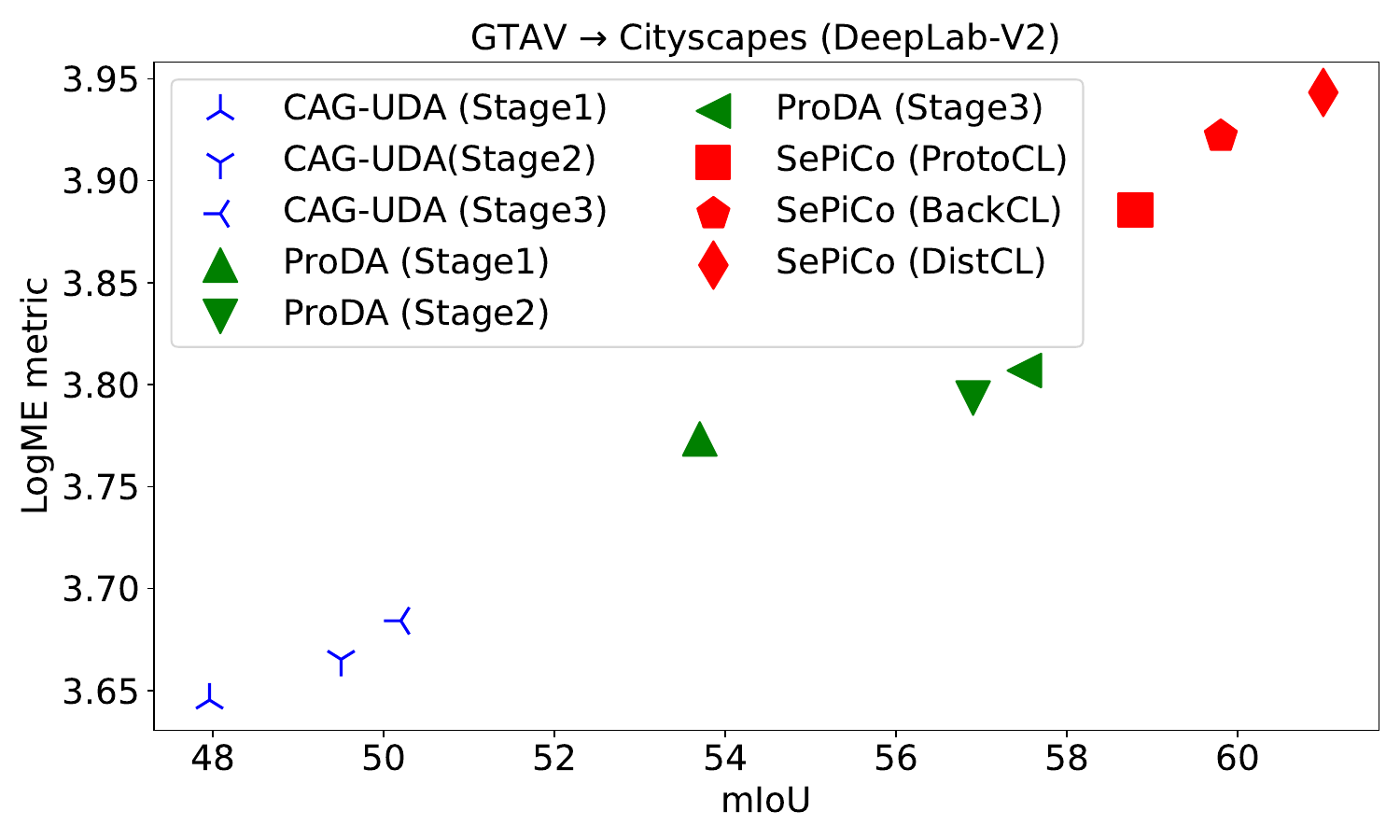}
    \VspaceBefore
    \caption{Comparison results of model transferability for different methods trained on GTAV $\to$ Cityscapes.}
    \label{Fig_logme}
    \VspaceAfter
\end{figure}
%##################################################################################################

\subsubsection{Quality of Model Generalization}
\label{sec:model_quality}
To quantify the generalization of our \method, we adopt a transferability metric (i.e., LogME~\cite{logme_YouLWL21}) to accurately assess the transferability of the model trained on GTAV $\to$ Cityscapes to the target dataset. Specifically, LogME calculates the maximum value of label evidence given extracted features by the adapted models and can measure the quality of models. A model with a higher LogME value is likely to have good transfer performance. In the trial, we consider each pixel and its ground-truth label as a separate observation. Since using all observations of target data to calculate is too computationally expensive, we instead calculate the LogME metric within a single target image and then average them. Fig.~\ref{Fig_logme} shows the comparison results at different stages of CAG-UDA~\cite{zhang2019category} (Stage1, Stage2, and Stage3) and ProDA~\cite{ProDA_2021_CVPR} (Stage1, Stage2, and Stage3) and our one-stage pipeline \method (ProtoCL, BankCL, and DistCL). This case study confirms the strong generalization of our pixel contrast paradigm, which essentially learns a well-structured pixel embedding space by making full use of the prototype, bank, or distribution-aware semantic similarities from the source domain.

%##################################################################################################
\begin{table}[t]
    \centering
    \caption{Throughput measured for different networks (DeepLab-V2~\cite{chen2018deeplab} and SegF. Mit-B5~\cite{xie2021segformer}) on GTAV $\to$ Cityscapes. Results are obtained using V100-32G and throughput is measured using a batch size of 1.} \VspaceBefore
    \label{table:throughput}
    \resizebox{0.48\textwidth}{!}{
        \begin{tabular}{lccccc}
            \toprule[1.2pt]
            Method  & mIoU & \begin{tabular}[c]{@{}c@{}} Student \\ model \end{tabular} &  \begin{tabular}[c]{@{}c@{}} Teacher \\ model \end{tabular} &  \begin{tabular}[c]{@{}c@{}} ImageNet \\ pretrained model \end{tabular} & \begin{tabular}[c]{@{}c@{}} Training time \\ per iteration (s)\end{tabular} \\
            \hline
            Source Only~\cite{chen2018deeplab} & 38.6 & $\Image_s$ & - & - & 0.32 (1.00$\times$) \\
            DACS~\cite{DACS_2021_WACV}  & 52.1 & $\Image_s + \Image_t$ & $\Image_t$ & - & 1.16 (3.63$\times$) \\
            \bf \method (DistCL) & \bf 61.0 & $\Image_s + \Image_t$  & $\Image_s + \Image_t$  & - & 1.34 (4.18$\times$) \\
            \hline
            \hline
            Source Only~\cite{xie2021segformer}  & 44.5 & $\Image_s$ & - & - & 0.35 (1.00$\times$) \\
            DAFormer~\cite{lukas2021daformer}  & 68.3 & $\Image_s + \Image_t$  & $\Image_t$ & $\Image_s$ & 1.33 (3.80$\times$) \\
            \bf \method (DistCL) & \bf 70.3 & $\Image_s + \Image_t$  & $\Image_s + \Image_t$  & - & 1.45 (4.14$\times$) \\
            \bottomrule[1.2pt]
        \end{tabular}
    }
\end{table}
%##################################################################################################
\subsubsection{Throughput} \label{sec:throughput}
We first compare inputs of different baseline methods. Source Only~\cite{chen2018deeplab,xie2021segformer}, for instance, contains a student model and takes source ($\Image_s$) images as input. DACS~\cite{DACS_2021_WACV}, an advanced self-training method, introduces the teacher-student model. The student model takes both source ($\Image_s$) and target ($\Image_t$) images as input, and the target model takes target images as input to generate target pseudo labels. As for DAFormer~\cite{lukas2021daformer}, the state-of-the-art method based on a Transformer backbone, it also contains a teacher model and a student model. The training process is similar to DACS, except that DAFormer introduces an auxiliary ImageNet pre-trained model and takes source ($\Image_s$) images as input to distill knowledge from expressive thing features of ImageNet. In this work, we also contain a teacher model and a student model and introduce a very lightweight projection head into the network that generates a new pixel embedding space. And both teacher and student models take source ($\Image_s$) and target ($\Image_t$) images as input, as shown in Fig.~\ref{Fig_framework}. 

Then, we compute the throughput of the mentioned methods on the GTAV $\to$ Cityscapes task using different networks. Since the throughput of the test phase is the same with the same network, we compare the training time of one iteration in TABLE~\ref{table:throughput}. Compared to existing methods, we can observe that \method only introduces slight extra computation on either CNN-based or Transformer-based networks but significantly surpasses comparison methods.
%%%%%%%%%%%%%%%%%%%%%%%%%%%%%%%%%%%%%%%%%%%%%%%%%%%%%%%%%%%%%%%%%%%%%%%%%%%%%%%%%%%%%%%%%%%%%%%%%%%

%%%%%%%%%%%%%%%%%%%%%%%%%%%%%%%%%%%%%%%%%%%%%%%%%%%%%%%%%%%%%%%%%%%%%%%%%%%%%%%%%%%%%%%%%%%%%%%%%%%
% conclusion
\section{Conclusion}
\label{sec:conclusion}

In this paper, we present \method, a novel end-to-end adaptation framework tailored for semantic segmentation, which successfully enhances the potential of the self-training paradigm in conjunction with representation learning. 
Our main contribution is the discovery  of pixel contrast guided by different semantic concepts. Eventually, we propose a particular form of contrastive loss at the pixel level, which implicitly involves the joint learning of an infinite number of similar/dissimilar pixel pairs for each pixel representation of both domains. 
Additionally, we derive an upper bound on this formulation and transfer the originally intractable loss function into practical implementation. 
Though simple yet effective, it works surprisingly well. Extensive experiments demonstrate the superiority of \method on both daytime and nighttime segmentation benchmarks. 
%%%%%%%%%%%%%%%%%%%%%%%%%%%%%%%%%%%%%%%%%%%%%%%%%%%%%%%%%%%%%%%%%%%%%%%%%%%%%%%%%%%%%%%%%%%%%%%%%%%

\section*{ACKNOWLEDGMENTS}
This paper was supported by National Key R\&D Program of China (No. 2021YFB3301503), and also supported by the National Natural Science Foundation of China under Grant No. U21A20519.
\ifCLASSOPTIONcaptionsoff
  \newpage
\fi

\bibliography{reference}
\bibliographystyle{IEEEtran}

\vspace{-10mm}
\begin{IEEEbiography}[{\includegraphics[width=1in,height=1.25in,clip,keepaspectratio]{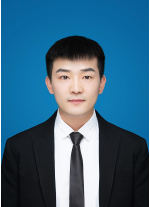}}]{Binhui Xie} is a Ph.D. student at the School of Computer Science and Technology, Beijing Institution of Technology. His research interests focus on computer vision and transfer learning.
\end{IEEEbiography}
\vspace{-10mm}

\begin{IEEEbiography}[{\includegraphics[width=1in,height=1.25in,clip,keepaspectratio]{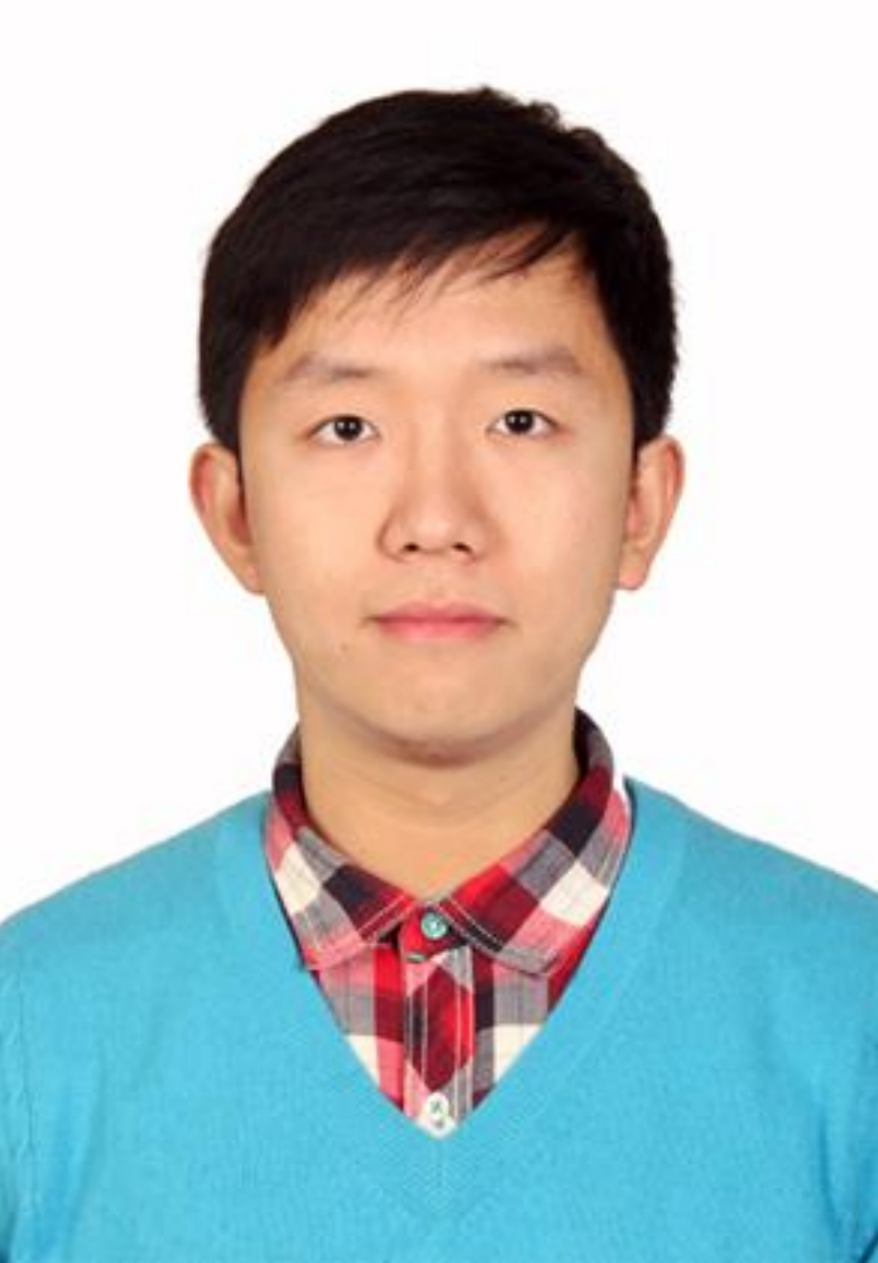}}]{Shuang Li} received the Ph.D. degree in control science and engineering from the Department of Automation, Tsinghua University, Beijing, China, in 2018.

He was a Visiting Research Scholar with the Department of Computer Science, Cornell University, Ithaca, NY, USA, from November 2015 to June 2016. He is currently an Associate Professor with the school of Computer Science and Technology, Beijing Institute of Technology, Beijing. His main research interests include machine learning and deep learning, especially in transfer learning and domain adaptation.
\end{IEEEbiography}
\vspace{-10mm}

\begin{IEEEbiography}[{\includegraphics[width=1in,height=1.25in,clip,keepaspectratio]{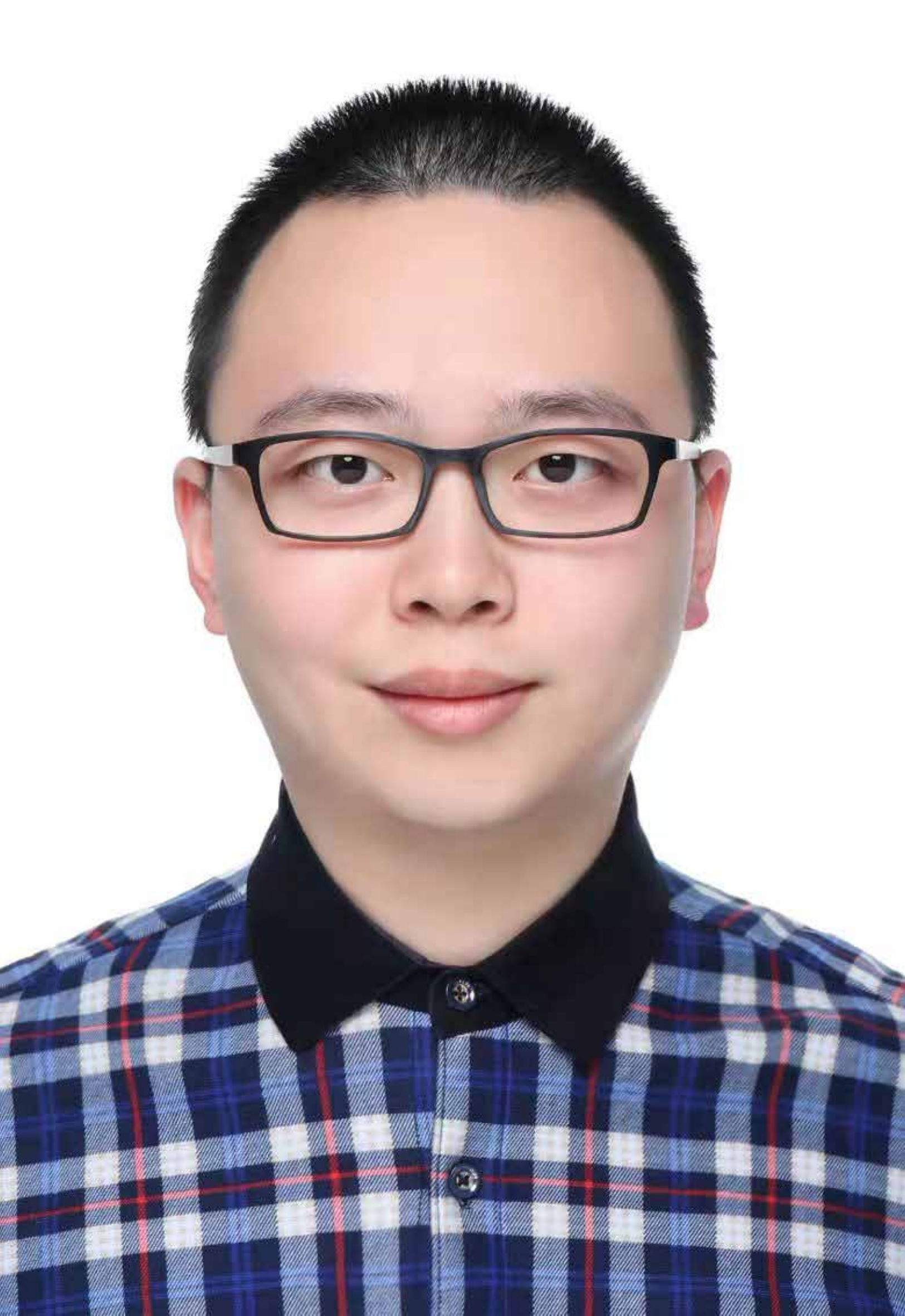}}]{Mingjia Li} is an undergraduate student at the School of Computer Science and Technology, Beijing Institution of Technology. His research interests focus on computer vision and transfer learning.
\end{IEEEbiography}
\vspace{-10mm}

\begin{IEEEbiography}[{\includegraphics[width=1in,height=1.25in,clip,keepaspectratio]{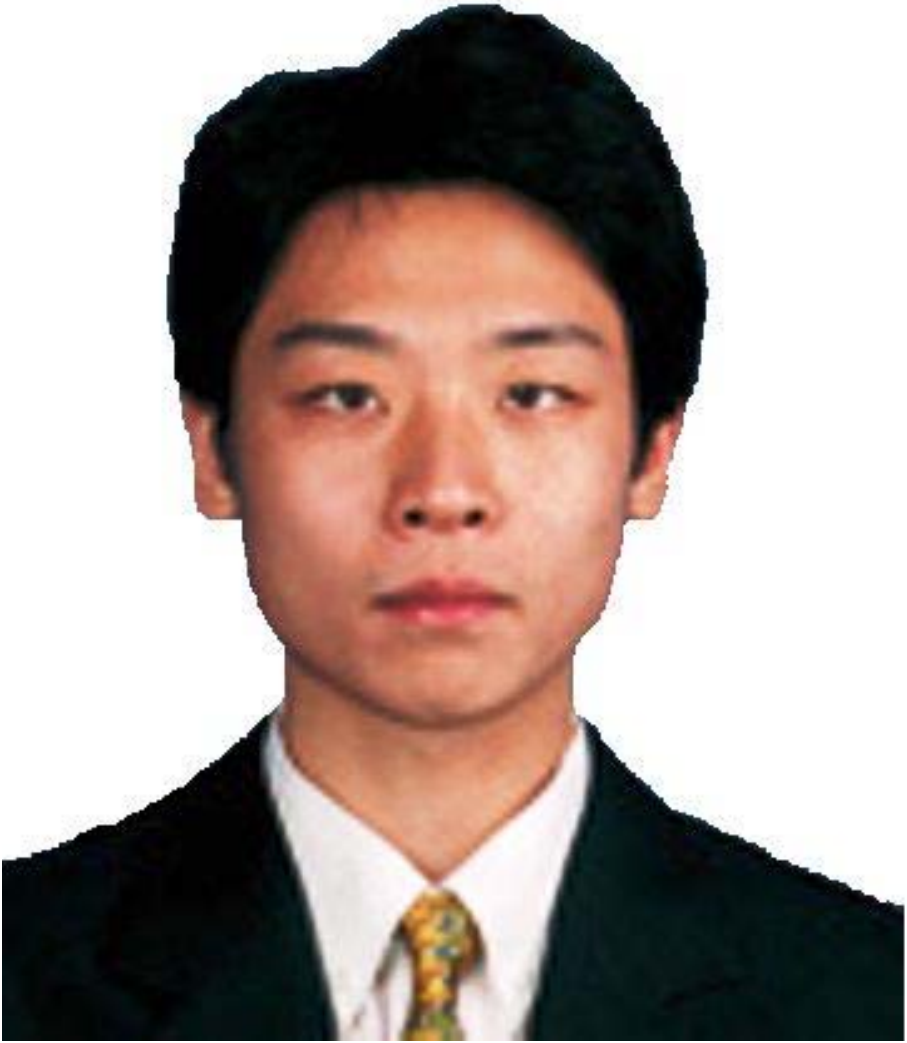}}]{Chi Harold Liu}(SM'15) receives a Ph.D. degree in Electronic Engineering from Imperial College, UK in 2010, and a B.Eng. degree in Electronic and Information Engineering from Tsinghua University, China in 2006.

  He is currently a Full Professor and Vice Dean at the School of Computer Science and Technology, Beijing Institute of Technology, China. Before moving to academia, he worked for IBM Research - China as a staff researcher and project manager from 2010 to 2013, worked as a postdoctoral researcher at Deutsche Telekom Laboratories, Germany in 2010, and as a Research Staff Member at IBM T. J. Watson Research Center, USA in 2009. His current research interests include the big data analytics, mobile computing, and machine learning. He received the IBM First Plateau Invention Achievement Award in 2012, ACM SigKDD'21 Best Paper Runner-up Award, and IEEE DataCom'16 Best Paper Award. He has published more than 100 prestigious conference and journal papers and owned 26 EU/UK/US/Germany/Spain/China patents. He serves as the Associate Editor for IEEE TRANSACTIONS ON NETWORK SCIENCE AND ENGINEERING, Area Editor for KSII Trans. on Internet and Information Systems, the Symposium Chair for IEEE ICC 2020 on Next Generation Networking, and served as the (Lead) Guest Editor for IEEE Transactions on Emerging Topics in Computing and IEEE Sensors Journal. He was the book editor for 11 books published by Taylor \& Francis Group, USA and China Machine Press, China. He also has served as the general chair of IEEE SECON'13 workshop on IoT Networking and Control, IEEE WCNC'12 workshop on IoT Enabling Technologies, and ACM UbiComp'11 Workshop on Networking and Object Memories for IoT. He was a consultant to Asian Development Bank, Bain \& Company, and KPMG, USA, and the peer reviewer for Qatar National Research Foundation, National Science Foundation, China, Ministry of Education and Ministry of Science and Technology, China. He is a senior member of IEEE and a Fellow of IET, British Computer Society, and Royal Society of Arts.
\end{IEEEbiography}
\vspace{-10mm}

\begin{IEEEbiography}[{\includegraphics[width=1in,height=1.25in,clip,keepaspectratio]{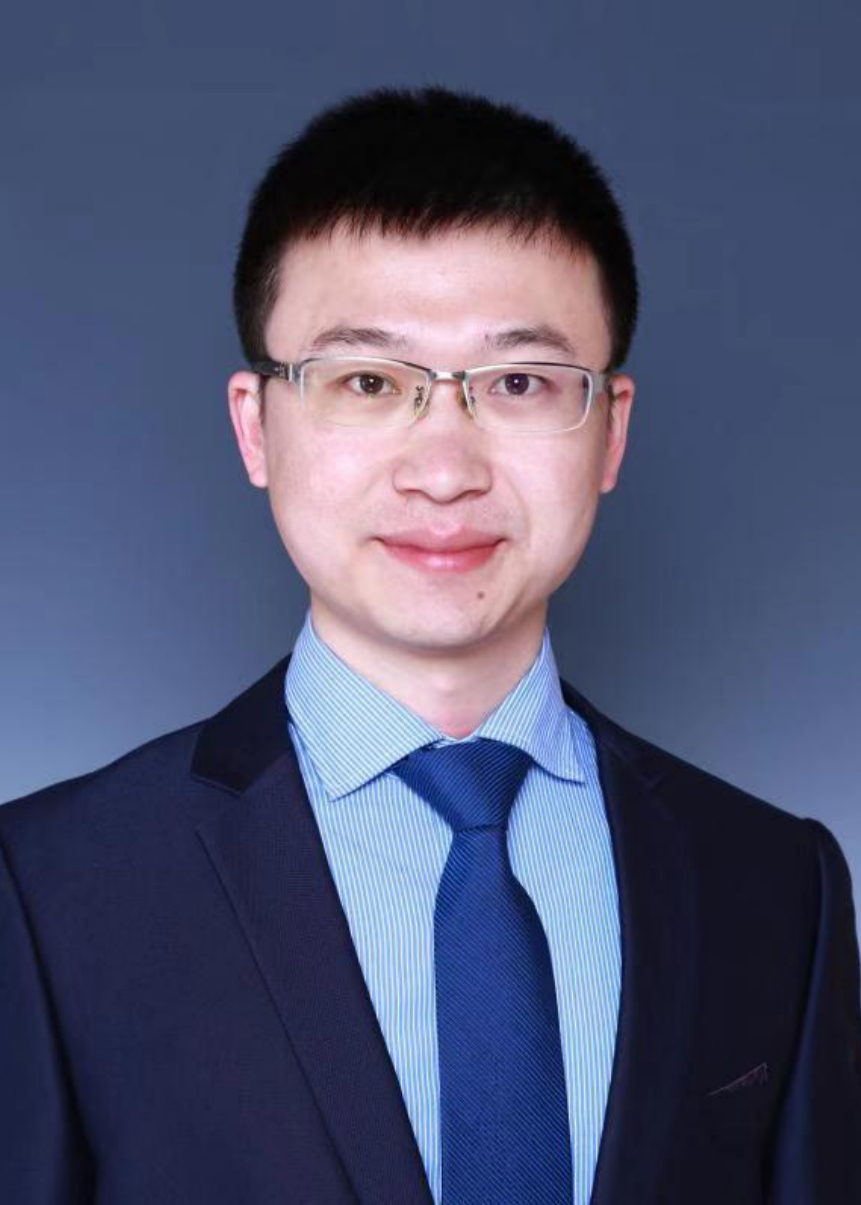}}]
  {Gao Huang} is an Associate Professor in the Department of Automation, Tsinghua University. He was a Postdoctoral Researcher in the Department of Computer Science at Cornell University. He received the PhD degree in Control Science and Engineering from Tsinghua University in 2015, and B.Eng degree in Automation from Beihang University in 2009. He was a visiting student at Washington University at St. Louis and Nanyang Technological University in 2013 and 2014, respectively. His research interests include machine learning and computer vision.
\end{IEEEbiography}
\vspace{-10mm}

\begin{IEEEbiography}[{\includegraphics[width=1in,height=1.3in,clip,keepaspectratio]{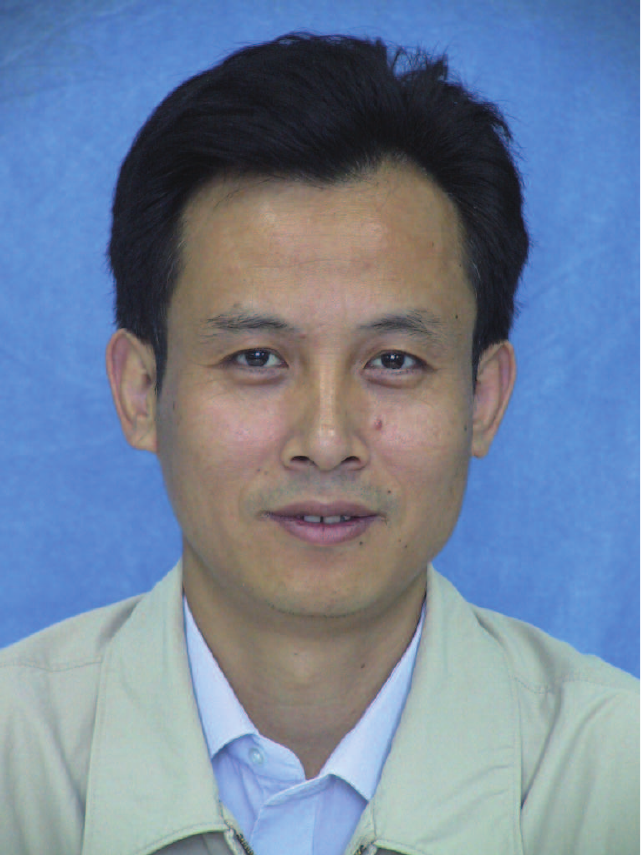}}]{Guoren Wang} received the BSc, MSc, and PhD degrees from the Department of Computer Science, Northeastern University, China, in 1988, 1991 and 1996, respectively. Currently, he is a Professor and the Dean with the School of Computer Science and Technology, Beijing Institute of Technology, Beijing, China. His research interests include XML data management, query processing and optimization, bioinformatics, high dimensional indexing, parallel database systems, and cloud data management. He has published more than 100 research papers.
\end{IEEEbiography}

\end{document}